\documentclass[letterpaper, 10 pt, conference]{ieeeconf}  

\IEEEoverridecommandlockouts                              

\usepackage{makeidx}  
\usepackage{etoolbox} 
\usepackage{graphicx}
\usepackage{svg}
\usepackage{amsmath}
\usepackage{algorithm}
\usepackage[noend]{algpseudocode}
\usepackage{svg}
\usepackage{subcaption}
\usepackage{cite}
\usepackage{booktabs}
\usepackage{tikz}
\usepackage{caption}
\usepackage[margin=0.75in]{geometry}

\usepackage{amsthm}
\usepackage{cite}
\usepackage{comment}

\usepackage{fancyhdr}
\fancypagestyle{withfooter}{
  
  \fancyhead[L]{}
  \fancyhead[R]{}
  \fancyfoot[C]{\footnotesize Presented at the 2025 IEEE ICRA Workshop on Field Robotics}
}

\title{\LARGE \bf 
End-to-End Framework for Robot Lawnmower Coverage Path Planning 
using Cellular Decomposition
}

\author{Shah Nikunj$^{1}$, Utsav Dey$^{1}$ and Kenji Nishimiya$^{1}$
\thanks{$^{1}$Honda R\&D Co., Ltd. Wako-shi, Japan \tt\small [shah\_nikunj, utsav\_dey, 
kenji\_nishimiya]@jp.honda}%
}

\begin{document}

\maketitle
\thispagestyle{withfooter}
\pagestyle{withfooter}
\begin{abstract}

 Efficient Coverage Path Planning (CPP) is necessary for autonomous robotic lawnmowers to effectively navigate and maintain lawns with diverse and irregular shapes. This paper introduces a comprehensive end-to-end pipeline for CPP, designed to convert user-defined boundaries on an aerial map into optimized coverage paths seamlessly. The pipeline includes user input extraction, coordinate transformation, area decomposition and path generation using our novel AdaptiveDecompositionCPP algorithm, preview and customization through an interactive coverage path visualizer, and conversion to actionable GPS waypoints. The AdaptiveDecompositionCPP algorithm combines cellular decomposition with an adaptive merging strategy to reduce non-mowing travel thereby enhancing operational efficiency. Experimental evaluations, encompassing both simulations and real-world lawnmower tests, demonstrate the effectiveness of the framework in coverage completeness and mowing efficiency.

\end{abstract}


%

\section{Introduction}
\label{sec:introduction}

Coverage Path Planning (CPP) in mobile robotics aims to generate paths that guarantee every point within a specified region is traversed \cite{howie_choset_2001, Galceran2013ASO}. In the context of robotic lawnmower applications, irregular lawn shapes and obstacles complicate the task of achieving total coverage and a visually appealing cut. Many existing CPP methods \cite{Choset1998CoveragePP, alex_2007, Luo2002AST} focus on achieving high coverage but give less attention to operational efficiency and lawn aesthetics. Operational efficiency here involves minimizing unnecessary turning maneuvers and nonmowing travel; aesthetics requires producing straight, uniform cuts with minimal deviations and overlaps \cite{agronomy11122567}.

Achieving both efficiency and aesthetics calls for smooth, continuous paths that reduce abrupt turns and idle maneuvers. In real-world lawn mowing, operators often rely on user-defined boundaries and adjust for varying mower widths and cutting parameters.  Unfortunately to the best of our knowledge, few existing methods integrate all these considerations (coverage, efficiency, aesthetics) into a single, adaptable pipeline. 

\begin{figure}
    \centering
    \includegraphics[width=\linewidth]{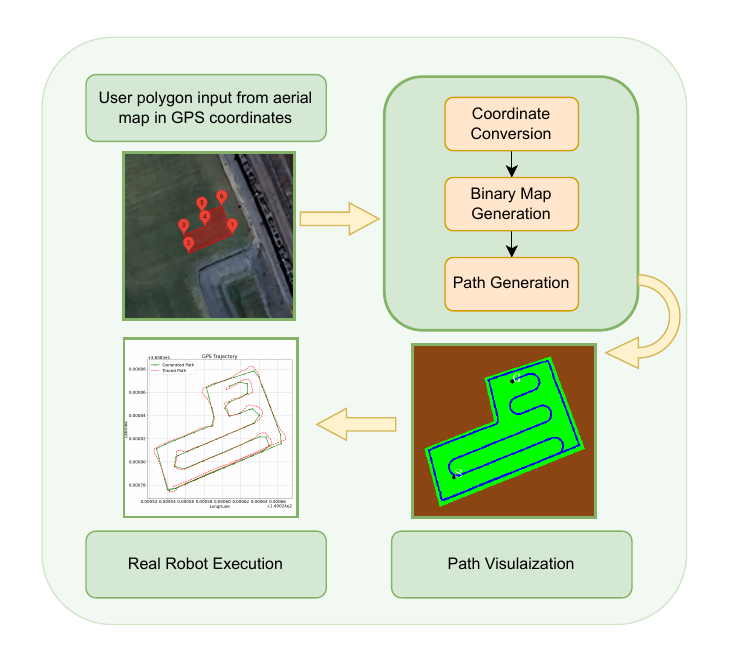}
    \caption{Proposed end-to-end framework execution of a sample lawn area with an actual lawnmower system. Clockwise from top: User-defined polygon input by marking GPS corner points on an aerial map; Input pre-processing pipeline; Visualization of the generated path using the preview simulator; GPS Trajectory comparison during autonomous path following.}
    
    \label{fig:l0}
    
\end{figure}

To address this gap, we introduce a modular framework for coverage path planning (CPP)—the AdaptiveDecompositionCPP algorithm. Our approach is designed to accommodate different cutting parameters while remaining flexible with regard to path-planning algorithms. It accounts for user-defined boundaries, mower geometry, and operational constraints. Fig.\ref{fig:l0} demonstrates our overall pipeline starting from user-drawn boundaries on an aerial map to GPS-based path execution on an actual lawnmower. We evaluated our method in both simulation and on real hardware. The results show that our approach achieves high coverage, reduces unnecessary turns and nonmowing travel, and also improves the overall aesthetics of the cut lawn.

In the following sections, we review related CPP methods (Section \ref{sec:related}), describe our adaptive algorithm and pipeline (Section \ref{sec:method}), and present experimental results that validate our approach. We conclude by discussing how future research may further enhance operational efficiency and cut quality.

\section{Related Work}
\label{sec:related}

Coverage Path Planning (CPP) is often compared to the well-known Traveling Salesman Problem (TSP) \cite{Modares2017UB-ANC, Kyaw2020Coverage}. The Covering Salesman Problem (CSP) \cite{10.1016/0166-218X(94)90008-6}, an adaptation of TSP where an agent is required to cover the vicinity of each city rather than just visiting it, is more closely related to the CPP task. By definition, CPP requires an agent to cover every point in a given area at least once.\cite{gabriely2001spanning} For instance, in the case of lawnmowers, this translates to mowing every part of the lawn. The TSP is known to be NP-hard, meaning it becomes computationally intensive to solve as the problem size increases, necessitating heuristic solutions \cite{Ouaarab2014Discrete}. Similarly, the task of mowing an entire grass-covered region, also known as the “lawnmower problem,” has also been established as NP-hard by Arkin et al. \cite{10.1016/S0925-7721(00)00015-8} Therefore, more practical approaches are often preferred over optimal ones in real-world applications.

Coverage Path Planning (CPP) is a relatively old problem, with some of the earliest algorithms being published in 1988 by Cao et al. \cite{Cao1988RegionFO} and Yasutomi et al. \cite{12333} Many current coverage path planning algorithms have been well summarized in the survey works by Galceran et al. \cite{Galceran2013ASO} and Choset et al. \cite{howie_choset_2001} Considering a 2D problem setup and a single agent in operation, there are broadly three types of approaches: cell decomposition methods, grid-based methods, and neural-based methods. \cite{Galceran2013ASO}

\begin{figure}
    \centering
    \begin{subfigure}[b]{0.31\linewidth}
        \centering
        \includegraphics[width=\linewidth]{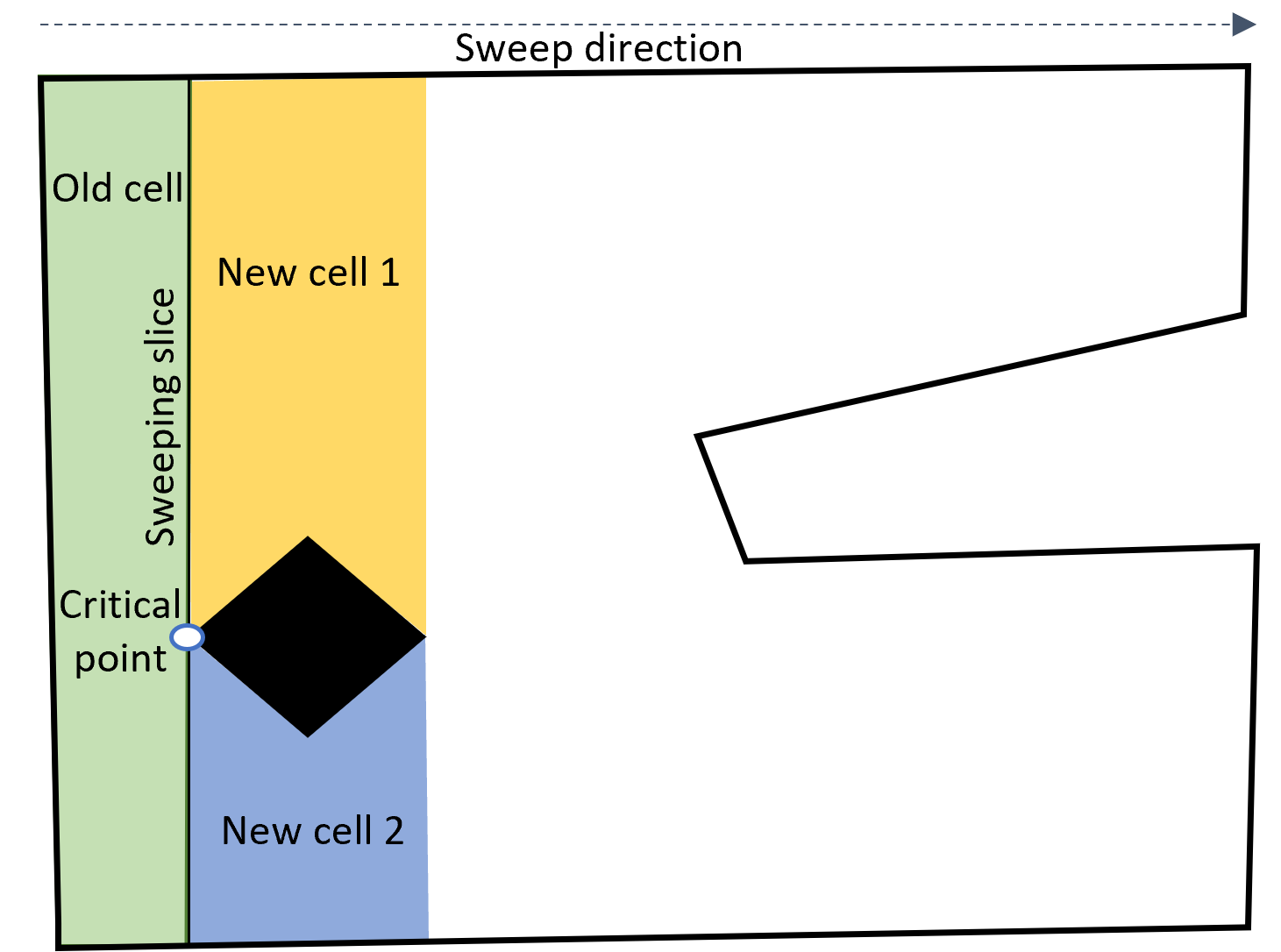}
        
        \label{fig:cell-dec1}
    \end{subfigure}
    \hfill
    \begin{subfigure}[b]{0.31\linewidth}
        \centering
        \includegraphics[width=\linewidth]{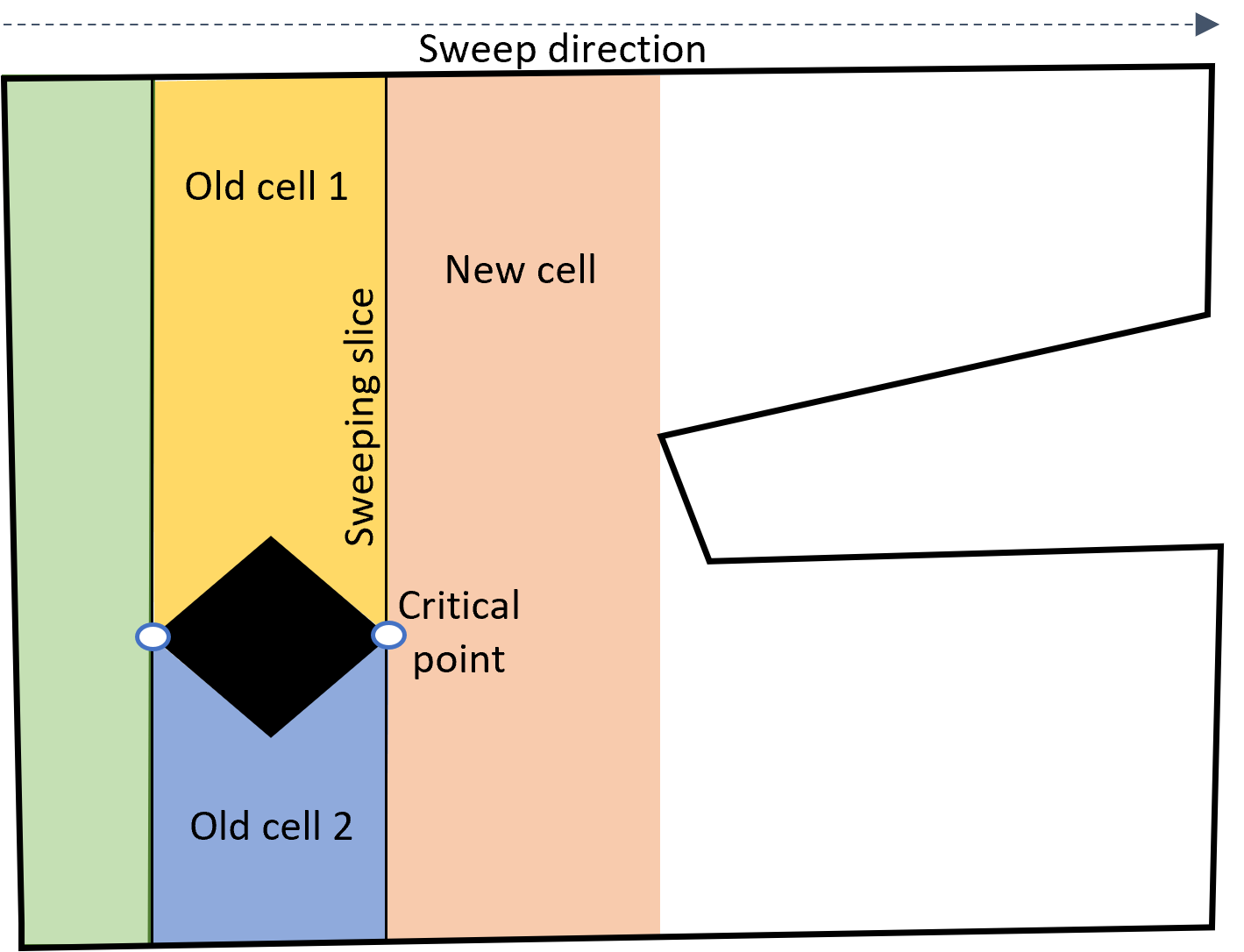}
        
        \label{fig:cell-dec2}
    \end{subfigure}
    \hfill
    \begin{subfigure}[b]{0.31\linewidth}
        \centering
        \includegraphics[width=\linewidth]{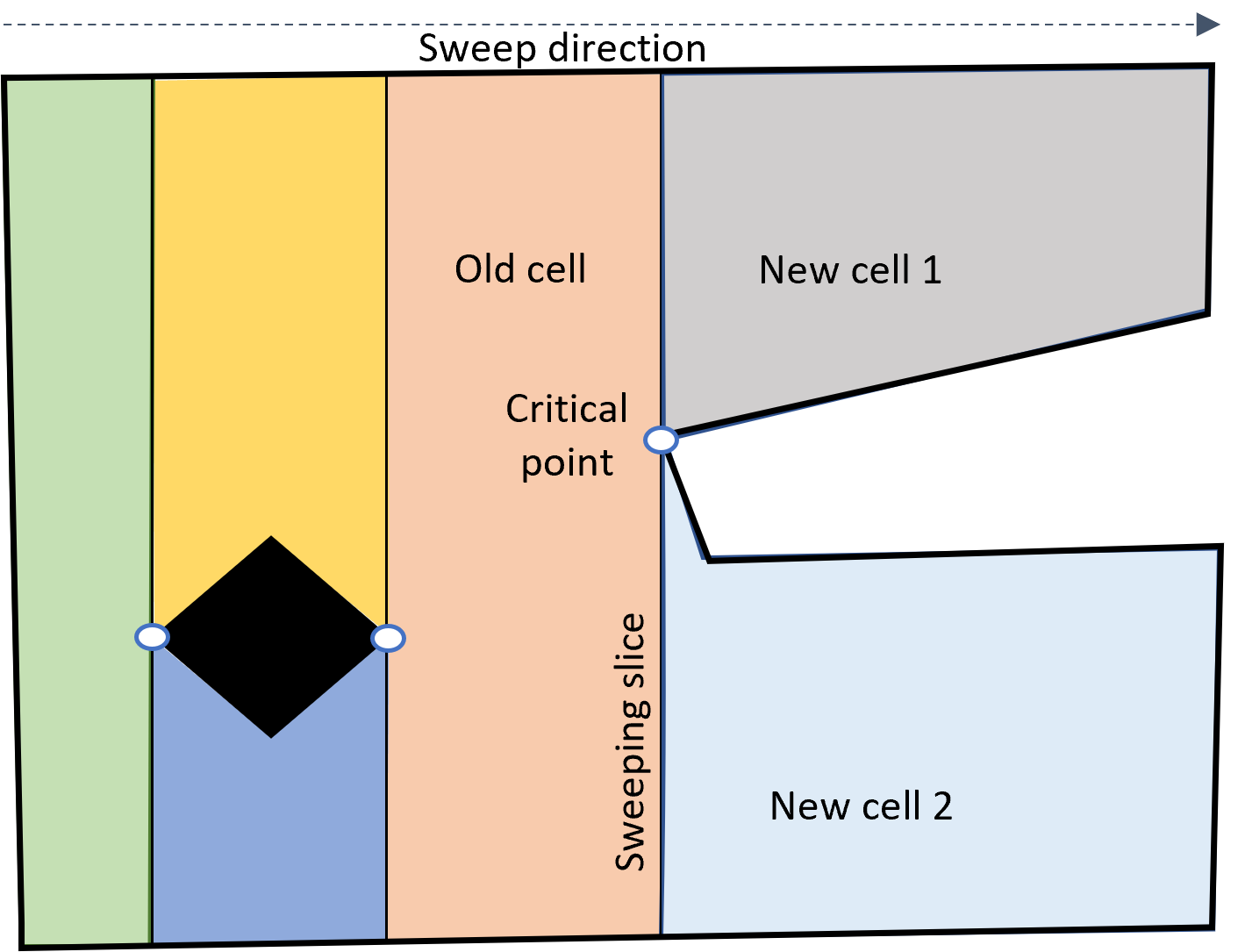}
        
        \label{fig:cel-dec3}
    \end{subfigure}
    \caption{Illustration of the boustrophedon cell decomposition method, as described in \cite{Choset1998CoveragePP}, dividing the area into simple subsections, incrementally from left to right.}
    \label{fig:l1}
\end{figure}

Cell decomposition methods such as \cite{Li2011Coverage, Brown2016The, Choset1998CoveragePP} aim to divide the free space into simpler, non-overlapping subsections. These subsections are free of obstacles and complex maneuvers and can usually be covered end-to-end by simple zigzag motion. One such example is the boustrophedon decomposition \cite{Choset1998CoveragePP},  where the decomposition lines are created by finding critical points around the obstacles (see Fig \ref{fig:l1}). Our algorithm is indeed a modified approach to this decomposition method. 

Grid-based methods divide the free space into small, equal-sized unit cells.\cite{alex_2007} The algorithm then plans the path from one grid cell to another nearby grid cell until all the eligible grid cells are covered. The technique used for planning may differ depending on the algorithm. For example, the work by Zelinsky et al. \cite{alex_2007} uses a distance transform that propagates a wavefront from the goal (assigned a value of 0) to the start, assigning a specific number to each grid element. Based on these distance values, the algorithm starts covering the cells from higher to lower values, resulting in a gradient-descent-like path.

Neural network-based approaches are the most recent development in this domain. They also divide the free space into grid cells and try to exploit neuron-like connection properties by considering each grid cell as a neuron. The works of Luo et al. \cite{Luo2002AST} and Guo et al. \cite{1641952} show the potential applications of neural-based approaches. These methods have the advantage of handling dynamic environments but are usually computationally heavy and may converge to the local optimum in some cases.

End-to-end approaches like ours are generally difficult to find in the literature. The most closely related work to our approach is a software tool developed by Hameed \cite{8311915}, which utilizes K-Means clustering and Dubins' curves for optimal Coverage Path Planning (CPP). Although the paper suggests that a robot equipped with a GPS could follow the generated path, it does not provide a comprehensive end-to-end implementation pipeline and is not tested on real hardware. The algorithm uses K-Means clustering to divide mowing tracks into sections; however, this method has several drawbacks. It requires users to manually determine the optimal number of clusters (K) based on starting coordinates, a process that is both challenging and prone to error, making it unsuitable for automated pipelines. Additionally, K-Means is vulnerable to becoming trapped in local optima depending on its initialization \cite{Yang2009An}. Moreover, the method does not optimize for tracks across all possible angles of decomposition, which limits its effectiveness in diverse scenarios.

Overall, while some approaches offer partial solutions to the CPP problem, there remains a lack of a comprehensive, adaptable, and computationally efficient end-to-end framework for diverse lawn shapes. Our proposed hybrid method fills this gap by combining cellular decomposition and adaptive merging techniques optimized for both coverage efficiency and practicality in real-world scenarios.
\section{Methodology}
\label{sec:method}
\subsection{Coverage Path Planning Algorithm}
\subsubsection{Overview and Notations}
\begin{figure}
    \centering
    \begin{subfigure}[b]{0.31\linewidth}
        \centering
        \includegraphics[width=\linewidth]{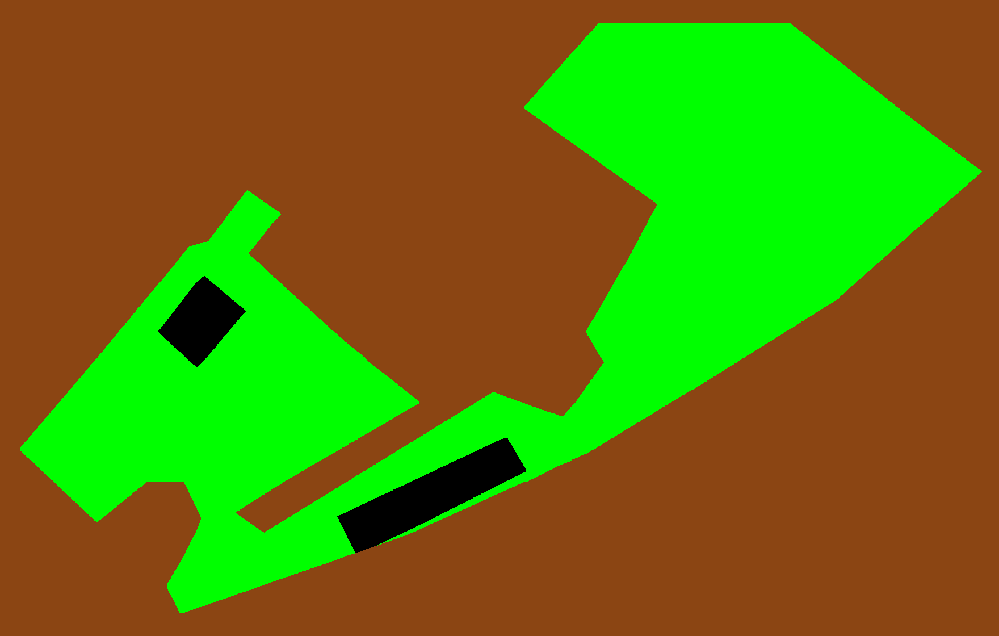}
        \caption{Field shape}
        \label{fig:field_shape}
    \end{subfigure}
    \hfill
    \begin{subfigure}[b]{0.31\linewidth}
        \centering
        \includegraphics[width=\linewidth]{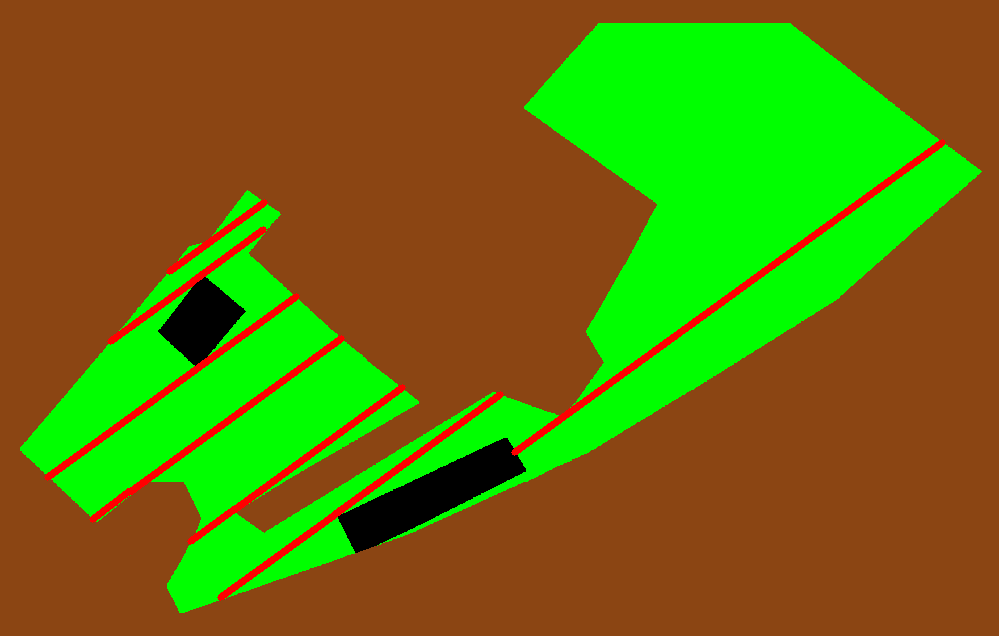}
        \caption{Decomposed}
        \label{fig:decomposed_sections}
    \end{subfigure}
    \hfill
    \begin{subfigure}[b]{0.31\linewidth}
        \centering
        \includegraphics[width=\linewidth]{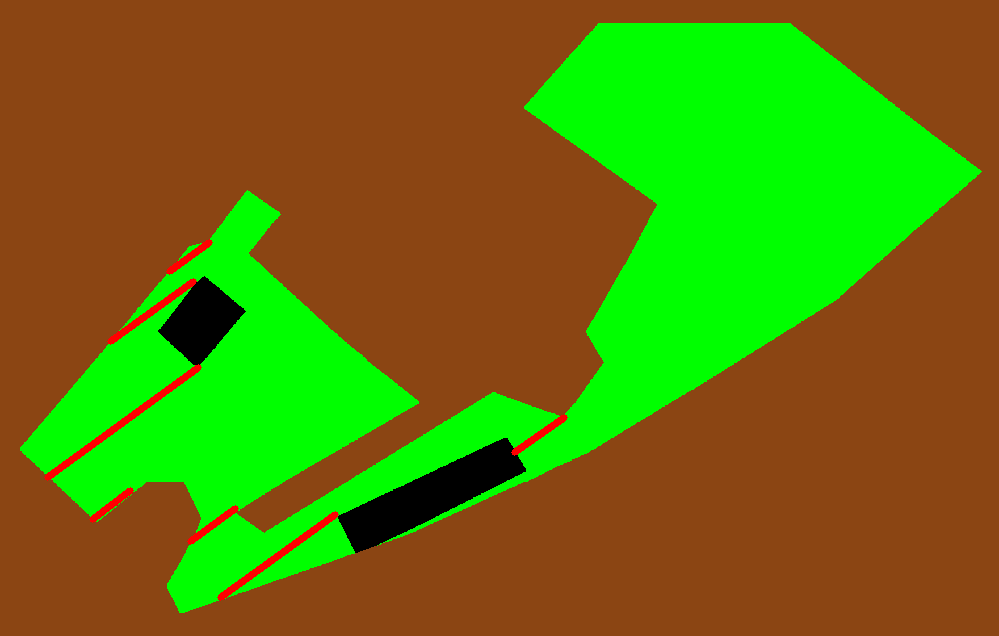}
        \caption{Merged}
        \label{fig:merged_sections}
    \end{subfigure}
    \caption{(a) The initial lawn shape provided as input. (b) The lawn is divided into 12 sections using cellular decomposition with a $36^\circ$ sweeping direction. (c) The final output after merging the sections into 5 optimized regions.}
    \label{fig:l2}
\end{figure}

In the given algorithm, \( L \) denotes the input image to be decomposed with m rows and n columns. \( D \) represents a decomposition of \( L \). \( |D| \) is the number of regions in that decomposition. \( \theta \) is an orientation angle ranging from \( 0^\circ \) to \( 180^\circ \). \( D^* \) and \( \theta^* \) are the optimal decomposition and its associated orientation. The goal is to minimize the number of regions. \( \mathcal{N}_{opt} \) is the set of all possible decompositions and their associated orientations. The procedure AdaptiveDecompositionCPP iterates over all possible orientations \(\theta\). It calls DecomposeMerge to obtain decompositions \( D_\theta \) at each angle. It then selects the decomposition with the fewest regions. Within DecomposeMerge, \( L_\theta \) is the image rotated by \( 90^\circ - \theta \); to align the decomposition lines along the image column. \( \mathcal{C} \) is the set of critical points across all iterations. \(\mathcal{C}_i\) is the ordered set of critical points for iteration \(i\) based on the sequence of insertion. \( \mathcal{R} \) is the set of regions drawn from the critical points. The procedure CriticalPoints identifies and collects critical points according to conditions defined later. \( L_{\theta_{i,j}} \) refers to the image region at row \( i \) and column \( j \). The procedure DrawLines updates the image with lines through these critical points. It forms the regions. The procedure Merge removes the first element from the ordered set \( \mathcal{C}_i \). Fig. \ref{fig:l2} shows the input lawn image, its decomposition into sections, and the final merged regions.

\begin{algorithm}
\caption{AdaptiveDecompositionCPP Algorithm}\label{alg:combined}
\begin{algorithmic}[1]

\Procedure{AdaptiveDecompositionCPP}{}
    \State $\mathcal{N}_{opt} \gets \emptyset$
    \For{$\theta \in \{0, 1, \ldots, 180\}$}
        \State $D_\theta \gets \text{DecomposeMerge}(\theta, L)$
        \State $\mathcal{N}_{opt} \gets \mathcal{N}_{opt} \cup \{ (D_\theta, \theta) \}$
    \EndFor
    \State $(D^*, \theta^*) \gets \text{argmin}_{(D, \theta)} \left\{ |D| : (D, \theta) \in \mathcal{N}_{opt} \right\}$
    \State \textbf{return} $D^*$
\EndProcedure

\Procedure{DecomposeMerge}{$\theta, L$}
    \State $L_\theta \gets \text{rotate}(L, 90^\circ - \theta)$
    \State $\mathcal{C} \gets \emptyset$
    \For{$i \in \{ 2, \ldots, n\}$}
        \State $\mathcal{C}_{i} \gets \text{CriticalPoints}(i, L_{\theta})$
        \State $\mathcal{C}_{i} \gets \text{Merge}(\mathcal{C}_{i})$
        \State $\mathcal{C} \gets \mathcal{C} \cup \mathcal{C}_{i}$
    \EndFor
    \State $\mathcal{R} \gets \text{DrawLines}(\mathcal{C}, L_{\theta})$
    \State \textbf{return} $\mathcal{R}$
\EndProcedure

\Procedure{CriticalPoints}{$i, L_{\theta}$}
    \State $\mathcal{C}_i \gets \langle \rangle$  \Comment{Initialize $\mathcal{C}_i$ as an ordered set}
    \For{$j \in \{ 1, \ldots, m\}$}
        \If{(region classification changes)}
            \If{(split or merge condition met)}
                \State $\mathcal{C}_i \gets \mathcal{C}_i \cup \langle (L_{\theta_{i-1,j}}, L_{\theta_{i,j}}) \rangle$
            \EndIf
        \EndIf
    \EndFor
    \State \textbf{return} $\mathcal{C}_i$
\EndProcedure

\Procedure{Merge}{$\mathcal{C}_i$}
    \State $\mathcal{C}_i \gets \mathcal{C}_i[2\ldots]$  \Comment{Remove the first element from the ordered set $\mathcal{C}_i$}
    \State \textbf{return} $\mathcal{C}_i$
\EndProcedure

\Procedure{DrawLines}{$\mathcal{C}, L_{\theta}$}
    \State $\mathcal{R} \gets \emptyset$
    \State $L_{\theta} \gets \text{updateImageLines}(\mathcal{C}, L_{\theta})$
    \State $\mathcal{R} \gets \text{extractRegions}(L_{\theta})$
    \State \textbf{return} $\mathcal{R}$
\EndProcedure

\end{algorithmic}
\end{algorithm}

\subsubsection{Conditions for a Point to be Critical}
Critical points are key locations where changes in the region's classification or connectivity occur, and they play a vital role in determining the structure of the decomposed regions. \newline
\textbf{Change in Region Classification:}
    A point is potentially critical if there is a change in its classification compared to the adjacent column. The current column is part of the lawn and the adjacent column is non-lawn and vice versa. If such a point is detected then the following conditions are verified to confirm it as a critical point:
\begin{itemize}

    \item \textbf{Splitting Condition:}
    A point is critical if it results in a previously connected region becoming disconnected in the next column. This can be visualized as a split in the free space.

    \item \textbf{Merging Condition:}
    A point is critical if it results in two previously disconnected regions becoming connected in the next column.
\end{itemize}



\subsubsection{Method}
The \texttt{AdaptiveDecompositionCPP} algorithm computes an optimal coverage path by iterating over angles \(\theta\) from 0 to 180 degrees (see Procedure AdaptiveDecompositionCPP, Line 3). For each angle, it calls DecomposeMerge to decompose the area (Line 4) and stores the result \(D_\theta\) with \(\theta\) in \(\mathcal{N}_{opt}\) (Line 5). It selects the decomposition \(D^*\) with the fewest regions to minimize turns (Line 6) and returns \(D^*\) (Line 7).

The DecomposeMerge procedure rotates the area \(L\) by \(90^\circ - \theta\) (Line 9) and initializes a set \(\mathcal{C}\) for critical points (Line 10). It iterates over columns \(i\) to find critical points using CriticalPoints (Line 12) and merges them using Merge (Line 13). The decomposition and merging process is illustrated in Fig. \ref{fig:l2}(b) and Fig. \ref{fig:l2}(c), where the lawn is initially decomposed into 12 sections (Fig. \ref{fig:l2}(b)) and subsequently merged into 5 optimized regions (Fig. \ref{fig:l2}(c)). The merged critical points are then used to draw paths with DrawLines (Line 15), which generates the boundaries or lines separating different regions based on the critical points identified and merged in previous steps. DrawLines first updates the image to reflect these lines (Line 29) and then extracts and returns the set of decomposed regions \(R\) (Line 30), providing the structured layout needed for optimal path planning. The result \(R\) is returned (Line 16).

The CriticalPoints procedure initializes an ordered set \(\mathcal{C}_i\) (Line 18). It iterates over each point in column \(i\) to detect changes in region classification (Line 20) and adds points to \(\mathcal{C}_i\) if a split or merge condition is met (Line 22). The set \(\mathcal{C}_i\) is returned (Line 23). If at least one critical point is detected, the start and end points of the lawn area in that specific column are also added to complete the set. The Merge procedure removes the first element from the ordered set \(\mathcal{C}_i\) (Line 25) and returns the modified set (Line 26).

\begin{figure}
    \centering
    \includegraphics[width=1.0\linewidth]{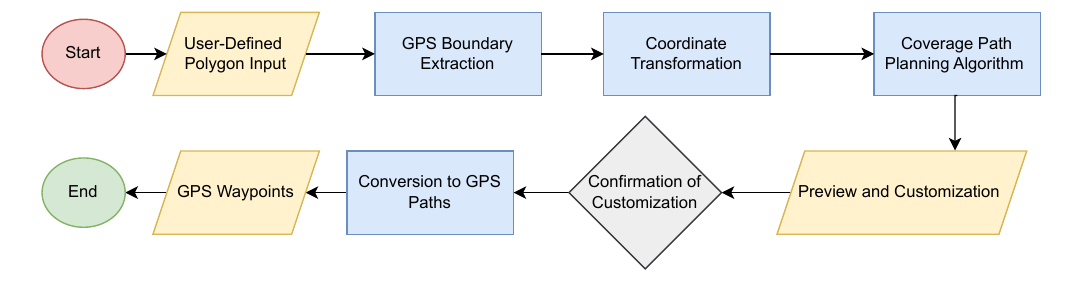}
    \caption{Flowchart illustrating the pipeline for generating optimized coverage paths, from user-defined polygon input to GPS waypoint output.}
    \label{fig:flowchart}
\end{figure}
\subsection{End-to-End Pipeline for Coverage Path Planning}

This section describes the end-to-end pipeline for converting GPS boundary data into coverage paths for autonomous systems. The pipeline is designed to achieve full coverage with minimal overlap and low redundancy in the path, as illustrated in the flow diagram in Fig. \ref{fig:flowchart}.

\subsubsection{User-defined Input and GPS Boundary Extraction}

The process begins with a user drawing a polygon on a map interface to define the boundary of the target area. The system then extracts the GPS coordinates of the vertices of this polygon. These act as the starting point for the subsequent steps in generating optimized coverage paths.

\subsubsection{Coordinate Transformation and Map Generation}

To simplify processing and visualization tasks, the extracted GPS coordinates are converted into a local coordinate system. This transformation makes it easier to handle computations that are otherwise complex in a geodetic coordinate system. Using the transformed coordinates, a map or grid representation of the area is generated, which serves as the input for the Coverage Path Planning Algorithm.

\subsubsection{Area Decomposition and Path Generation Using the Coverage Path Planning Algorithm}

This step utilizes the method described in the algorithm section for its robustness and computational simplicity. The Coverage Path Planning Algorithm manages both the decomposition of the area and the generation of coverage paths. This method is chosen for its balance of simplicity and effectiveness, but it can be replaced with another algorithm if different criteria or specific needs arise. The result of this step is a set of optimized paths represented in the local coordinate system.
\begin{figure}
 \centering
  \begin{tabular}{cc}
    \includegraphics[width=.445\linewidth]{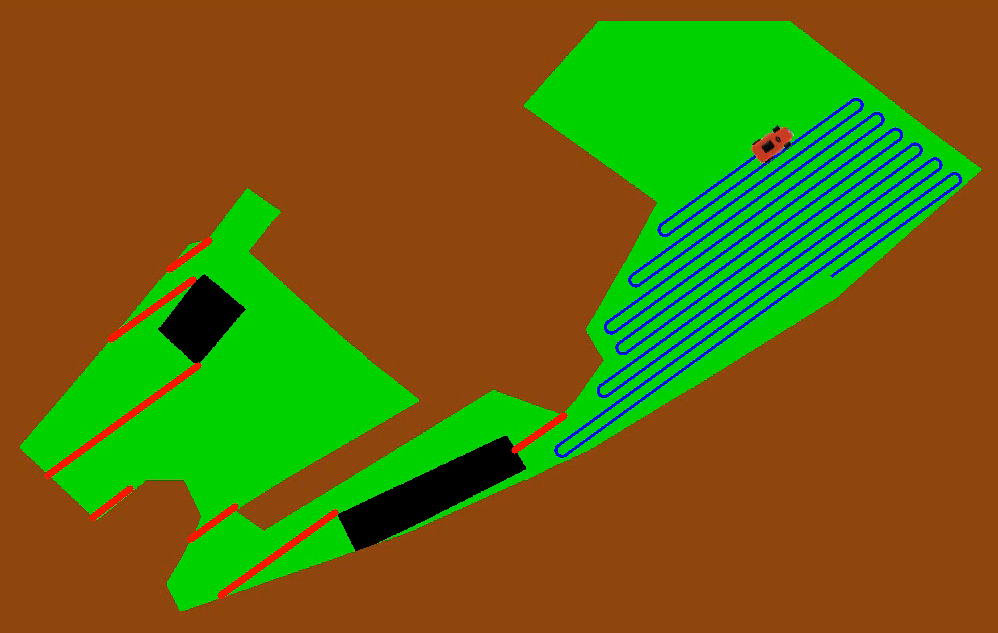} & \includegraphics[width=.445\linewidth]{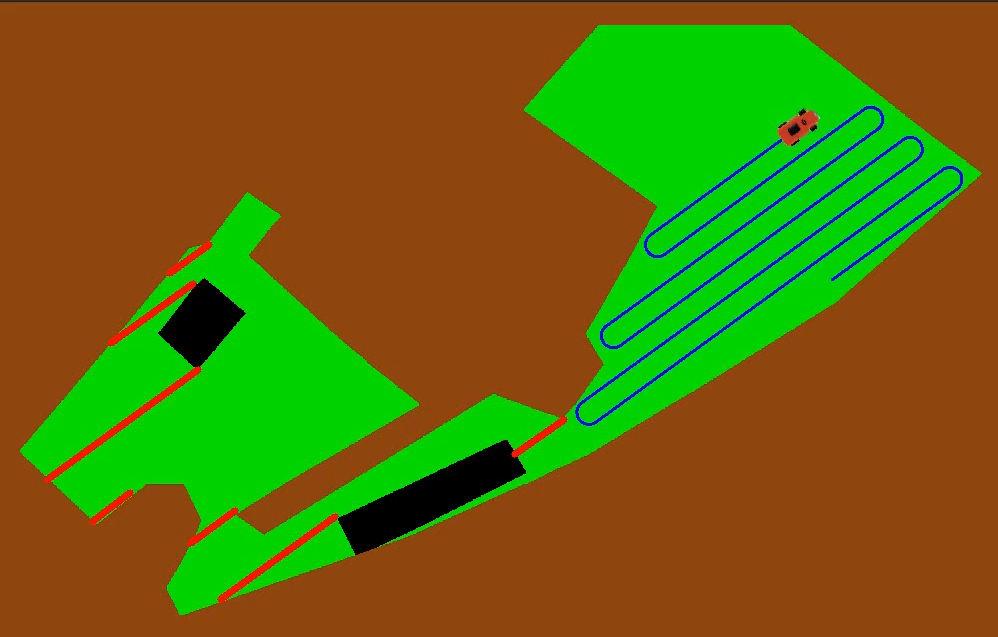} \\ \includegraphics[width=.445\linewidth]{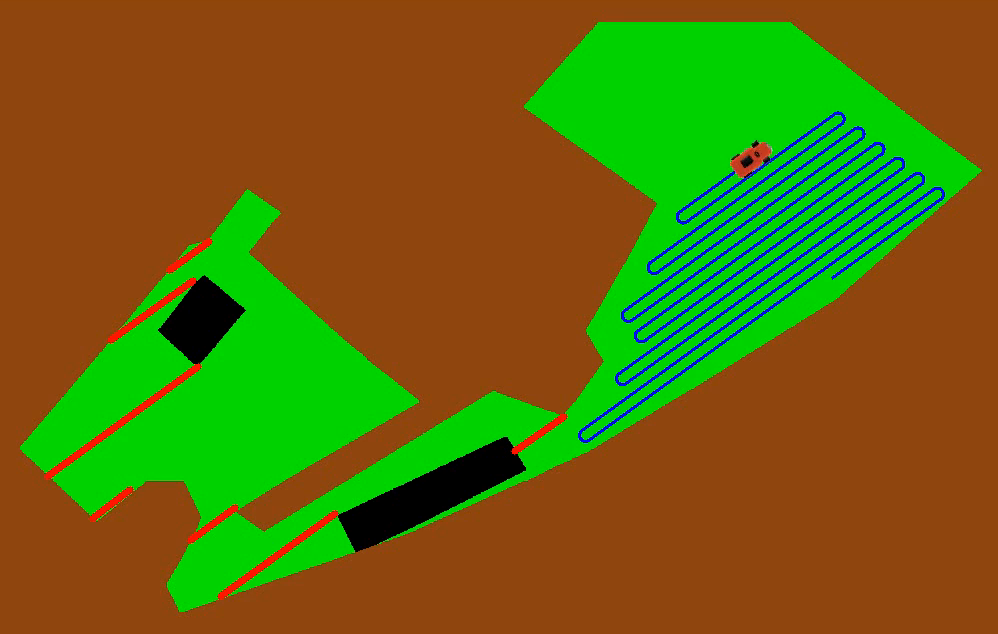} & \includegraphics[width=.445\linewidth]{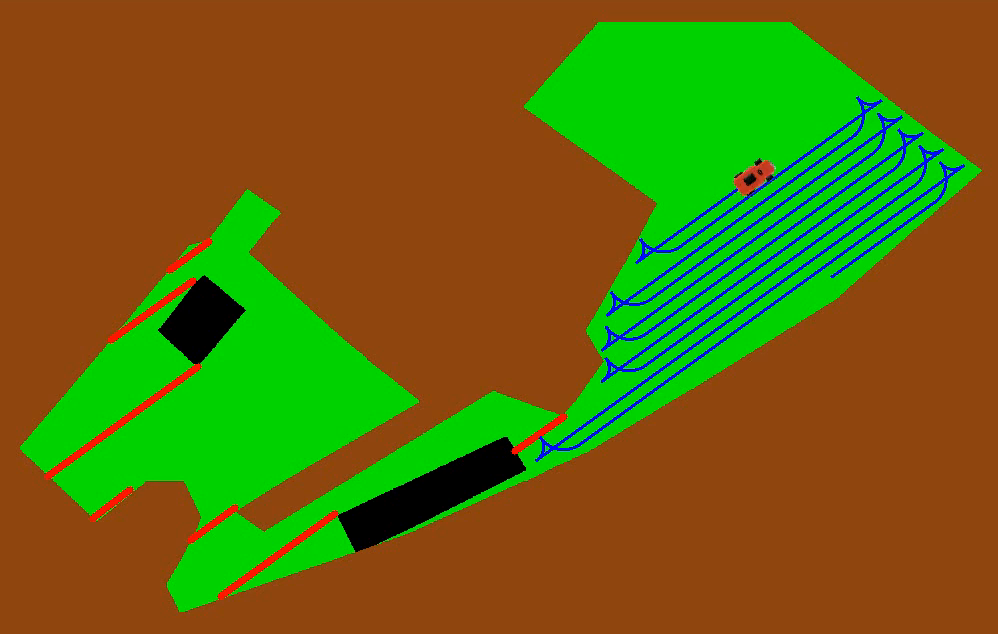} 
  \end{tabular}
  \caption{Comparison of different simulator configurations: Clockwise from the top left, the first image shows the default settings, the second image illustrates a high turn radius with an increased width, the third image depicts a greater distance from the boundary, and the fourth image shows the configuration when a U-turn is changed to a three-point turn.}
  \label{fig:simulator_different}
\end{figure}
\subsubsection{Interactive coverage path visualizer}

We developed a ROS-based coverage path visualizer to visualize, preview, and fine-tune the generated mowing trajectories prior to real-world execution. This environment renders the layout of the target lawn area and mowing paths, allowing users to adjust parameters such as the offset from the boundaries, the turning radius, and the mowing width in real-time.

 As shown in Fig. \ref{fig:simulator_different}, the first image displays the default settings, the second image illustrates a high turn radius with increased width, and the third image depicts a higher gap from the boundary, whereas the fourth image depicts a variation in turn type. Once satisfied with the configuration, they can proceed to the next step, which is generating the GPS output.

\subsubsection{Conversion to GPS Paths}

After the user confirms the path in the simulator, the optimized paths are converted back into GPS coordinates. This conversion makes the paths suitable for use by autonomous vehicles or mowers. The final output is a sequence of GPS waypoints that guide the autonomous system through the planned path, ensuring complete coverage of the area.

\section{Experiment}
\label{sec:experiments}

\subsection{Definitions}
\begin{itemize}
    \item \textbf{Coverage Percentage}: 
 It measures how fully the area is mowed and is critical to ensure full coverage without unmowed spots.
 \item \textbf{Non-Mowing Distance}: It represents the distance traveled when the mower is not actively cutting grass, such as moving between sections and minimizing it reduces wasted time and energy.
 \textbf{Distance per Coverage (DC)}: This metric is defined as ratio:
\[
 \frac{\text{Mowing Distance} + \text{Non Mowing Distance}}{\text{Coverage Percentage}}
\]
indicating how much distance is traveled per unit percentage of area covered. Lower values signify more efficient paths. 
\item \textbf{Mowing distance}: It is the distance traveled during mowing. 
 
\end{itemize}
\begin{table*}[ht]
    \centering
    \renewcommand{\arraystretch}{1} 
    \begin{tabular}{lcccc}
        \toprule
        & \textbf{TCD\cite{Galceran2013ASO}} & \textbf{BCD\cite{Choset2000CoverageOK}} & \textbf{Grid-Based\cite{Gabriely2002SpiralSTCAO}} & \textbf{Ours} \\
        \midrule
        Output Image &
        \includegraphics[width=0.15\textwidth]{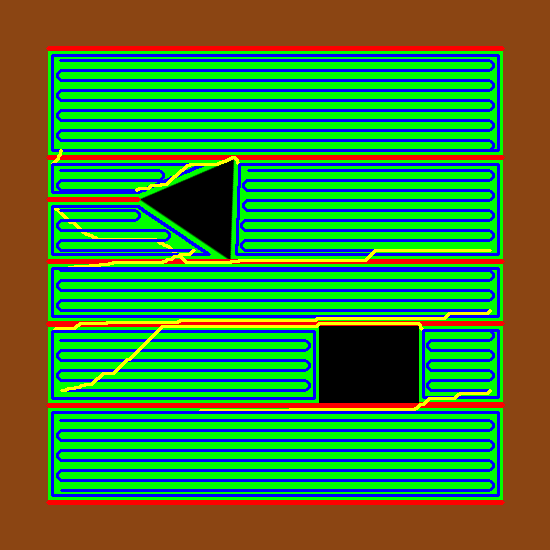} &
        \includegraphics[width=0.15\textwidth]{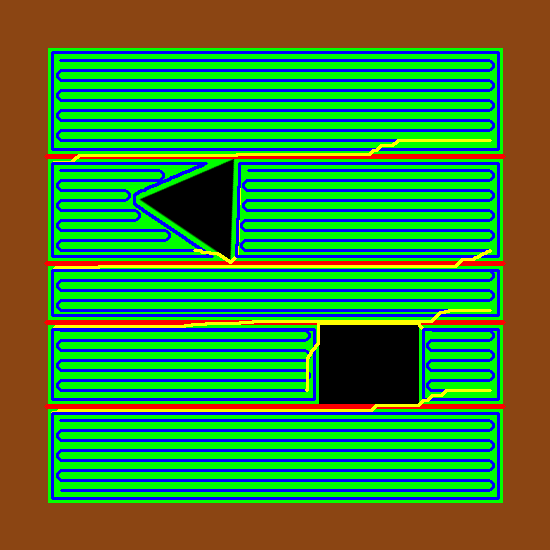} &
        \includegraphics[width=0.15\textwidth]{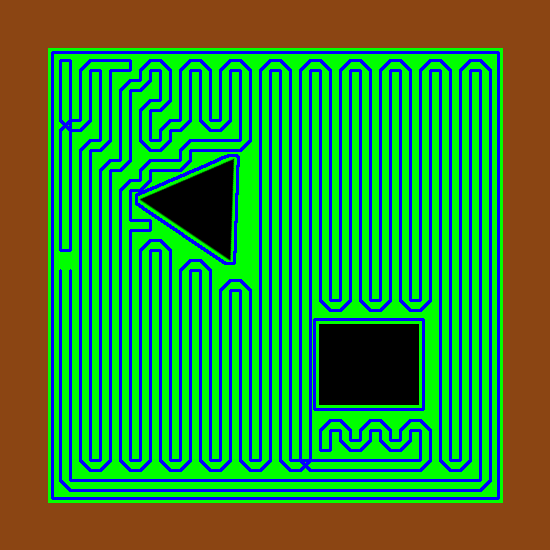} &
        \includegraphics[width=0.15\textwidth]{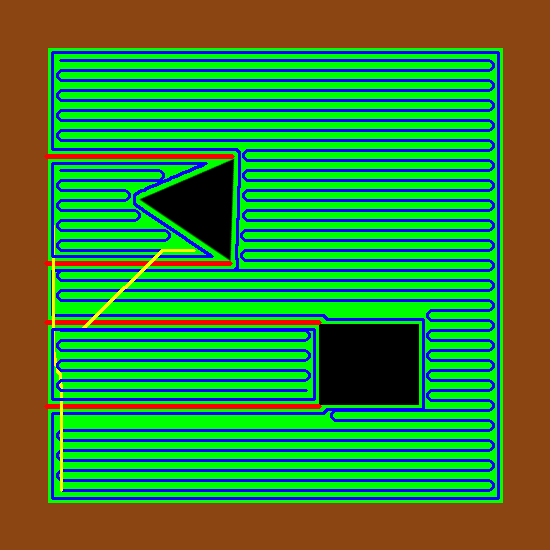} \\
        \midrule
        Number of Turns & \textbf{45} & 47 & 174 & 57 \\
        Mowing Distance & 20811 & 20608 & \textbf{19285} & 19856 \\
        Non Mowing Distance & 1946 & 2504 & - & \textbf{471} \\
        Coverage Percentage & 94.5\% & 96.8\% & 96.5\% & \textbf{97.2}\% \\
        Number of Decompositions & 8 & 7 & - & \textbf{3} \\
        Distance per Coverage & 240.9 & 238.7 & \textbf{199.8} & 208.9 \\
        \bottomrule
    \end{tabular}
    \caption{Comparison with baseline Coverage Path Planning algorithms (TCD stands for Trapezoidal Cell Decomposition and BCD stands for Boustrophedon Cell Decomposition). Yellow lines indicate the non-mowing path. Red lines represent subsection boundaries. Generated coverage paths are shown as blue lines. The base map is chosen so as to effectively visualize the differences between the compared approaches.}
    \label{tab:comparison}
\end{table*}

\subsection{Comparison with baseline algorithms}

We focus on deterministic methods based on geometry because they use the geometry of the lawn and do not rely on training data, making them easier to interpret, debug and refine. As discussed in the related work section, there are major varieties of non-learning based CPP algorithms. One is decomposition-based and the other is grid-based. Below is our observation from  Table \ref{tab:comparison}.

Our approach achieves better coverage than both Boustrophedon Decomposition \cite{Choset2000CoverageOK} and Trapezoidal Decomposition \cite{Galceran2013ASO} with lower mowing and non-mowing distance. We attribute the lower non-mowing distance to merging, which reduces the number of decompositions leading to fewer travels between decompositions. The lower mowing distance without any observable deterioration in the coverage percentage also signifies that there is comparatively less overlapping of paths in case of our approach.

The grid-based method \cite{Gabriely2002SpiralSTCAO} has slightly higher coverage percentage; however, the path planned by this method has a lot of abrupt turns in the middle which leads to a non-aesthetic cut of the lawn. Also, for practical use, turning is a more difficult and time-consuming operation for a mower, hence the very high number of turns is also a big disadvantage.

Given these observations, our approach proves to be more suitable for lawnmowing operations than existing methods. On the efficiency side, our lower nonmowing distance and comparable or better coverage percentages clearly indicate reduced unnecessary travel. From an aesthetic perspective, smoother and more direct paths, especially when compared to the numerous abrupt turns in the grid-based method, yield straighter lines and a more uniform cut. These measurable improvements align with our primary goals of maximizing efficiency and improving lawn appearance, as described in the introduction. Also, except for the grid-based approach (above 1000 ms) the average execution time of all other methods was noted to be below 100 ms on a desktop CPU system.

\subsection{Impact of Merging and Decomposition Angle on Key Metrics}

\begin{table*}[ht]
\centering
        \begin{tabular}{cccccc}
            \toprule
            & \multicolumn{2}{c}{Decomposition angle: reference} & \multicolumn{2}{c}{Decomposition angle: min sections} \\
            \cmidrule(r){2-3} \cmidrule(r){4-5}
            & Merging: No & Merging: Yes & Merging: No & Merging: Yes \\
            \midrule
            Results \#1 & \includegraphics[width=3.0cm]{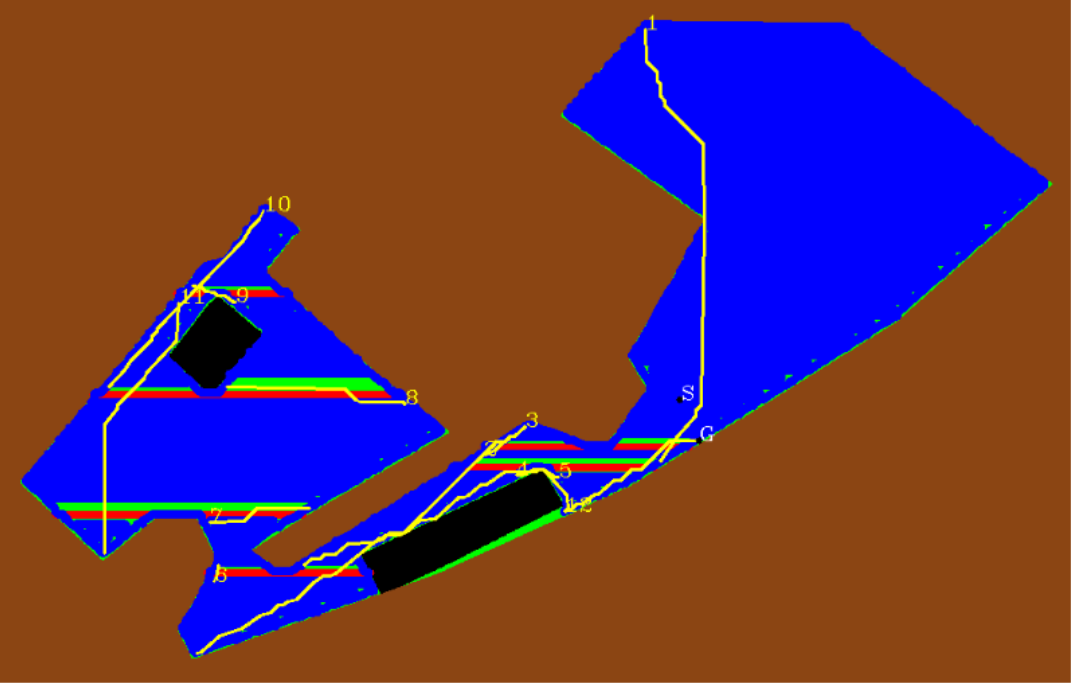} & \includegraphics[width=3.0cm]{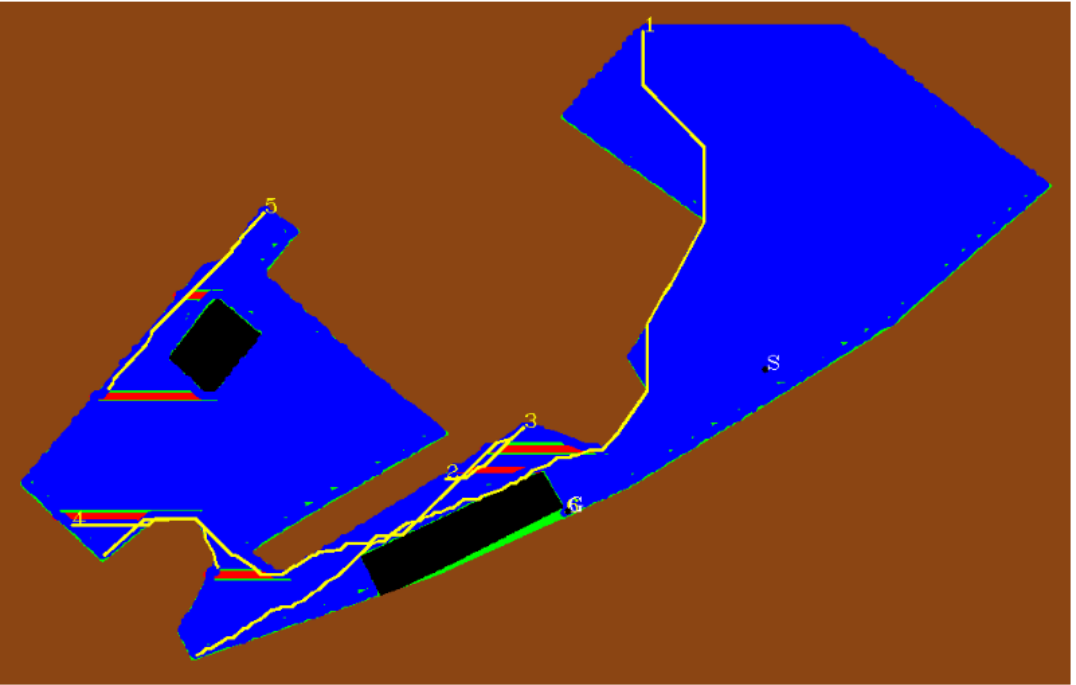} & \includegraphics[width=3.0cm]{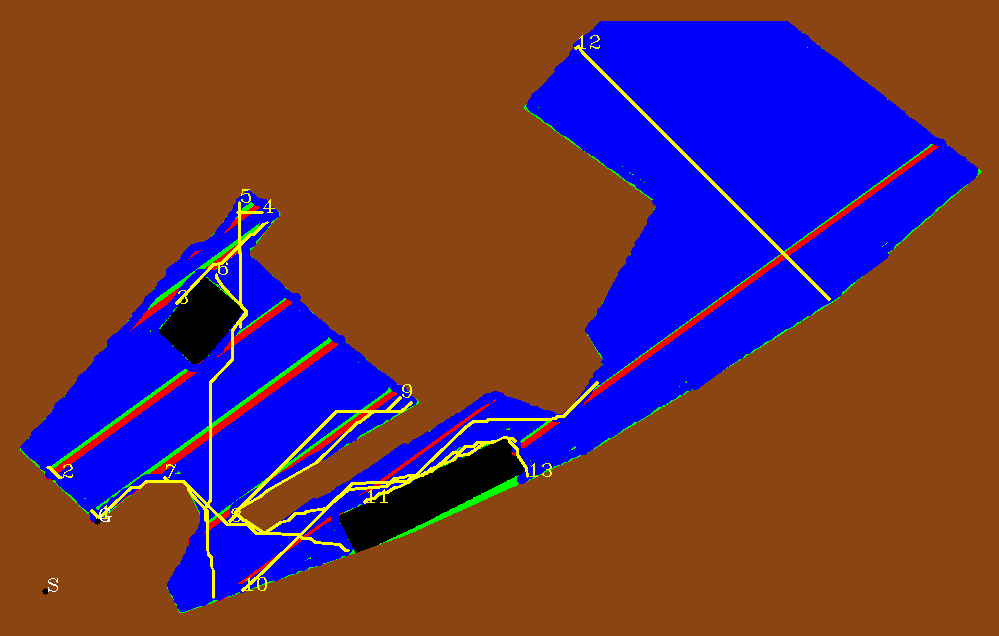} & \includegraphics[width=3.0cm]{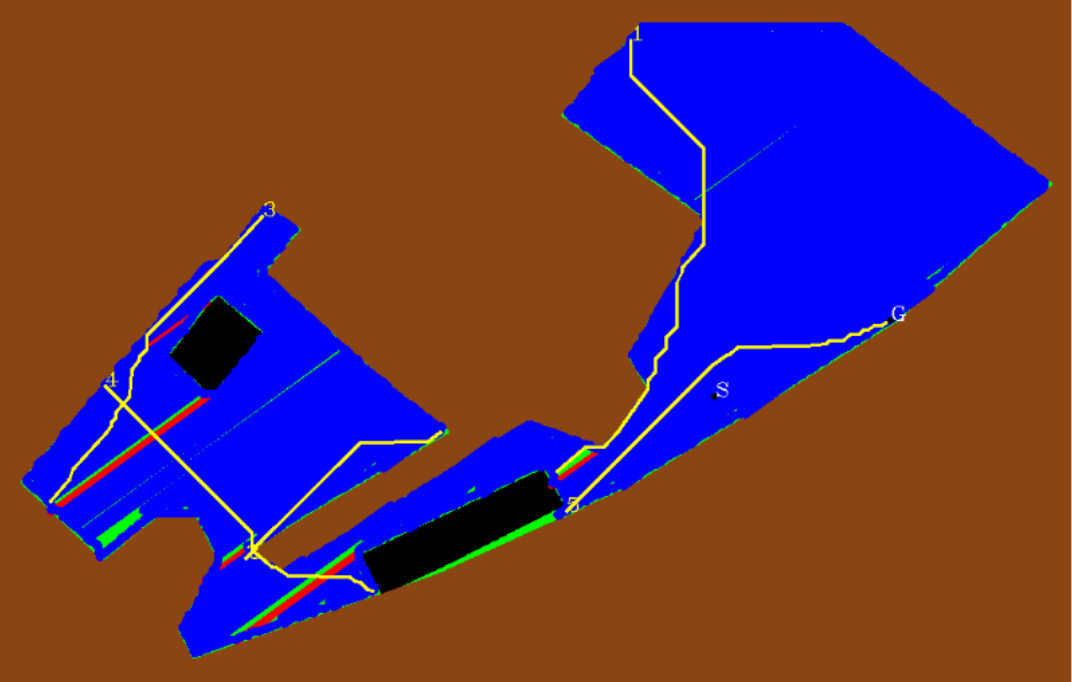} \\
            Angle & 0° & 0° & 36° & 36° \\
            Number of Decompositions & 12 & 6 & 12 & \textbf{5} \\
            Mowing Distance & 1904.6m & 1996.4m & \textbf{1798.7m} & 1902.6m \\
            Non Mowing Distance & 122.0m & 114.5m & 125.0m & \textbf{109.1m} \\
            Coverage \% & 95.18 & \textbf{97.61} & 96.39 & 97.35 \\
            Distance per Coverage & 21.27 & 21.62 & \textbf{20.03} & 20.65 \\
            \midrule
            Results \#2 & \includegraphics[width=2.5cm]{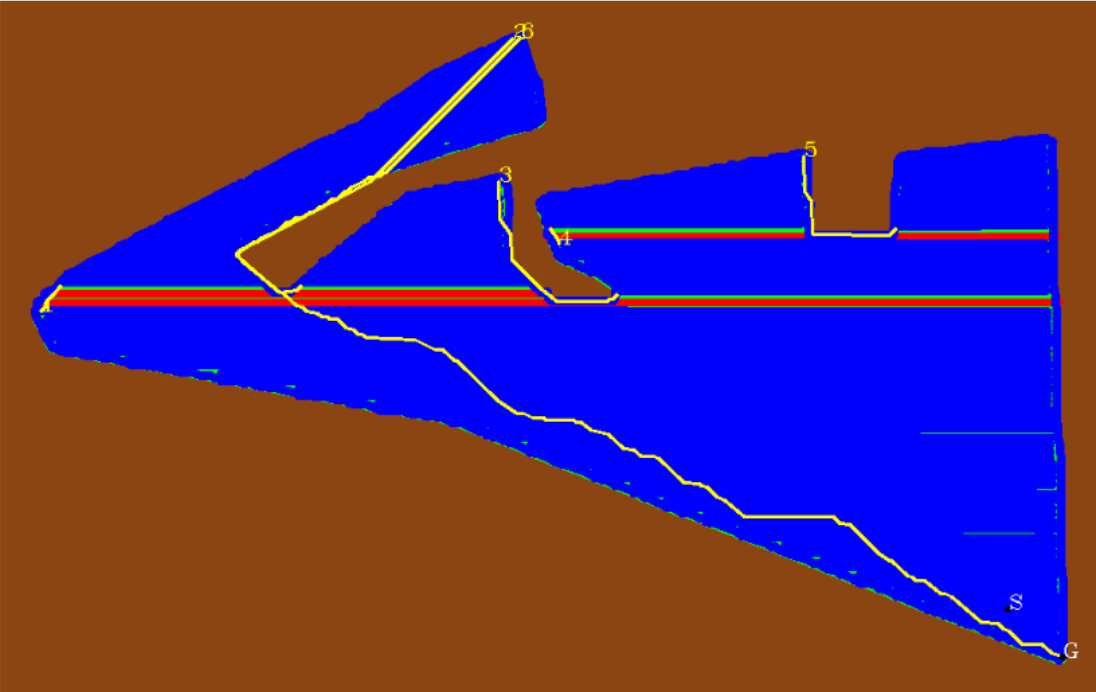} & \includegraphics[width=3.0cm]{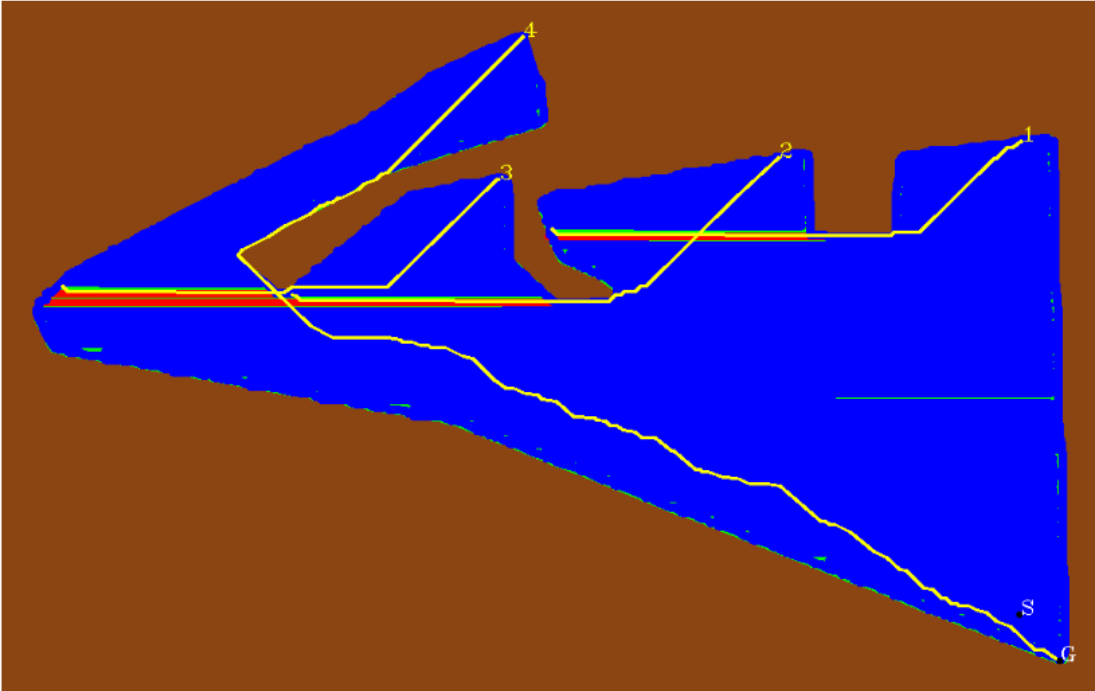} & \includegraphics[width=3.0cm]{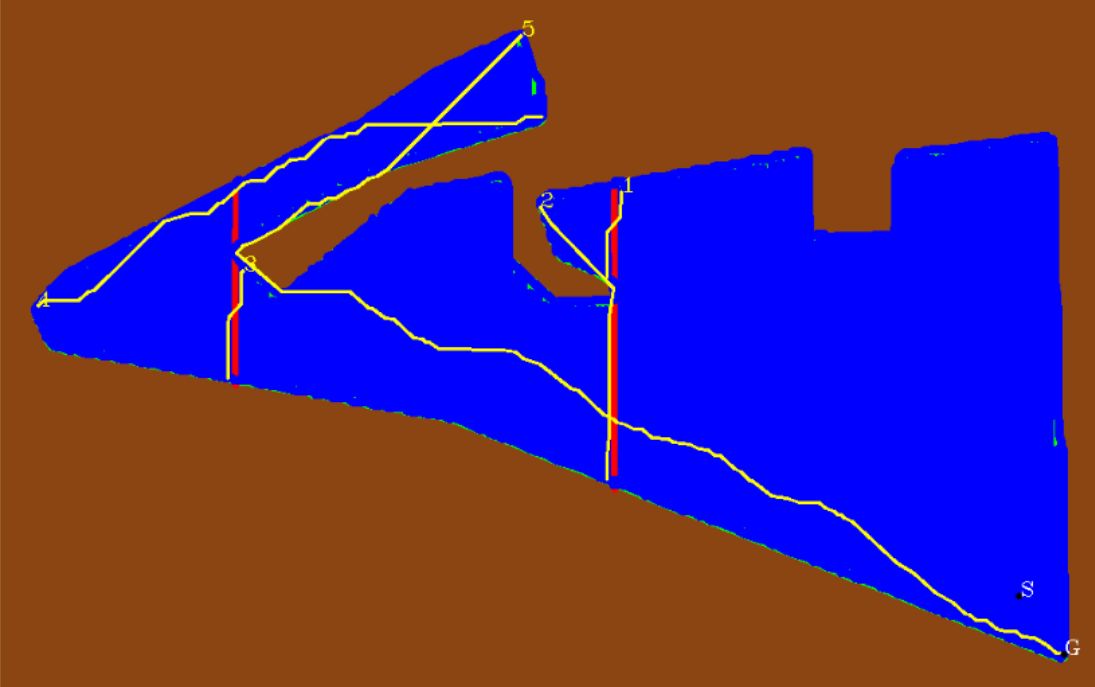} & \includegraphics[width=3.0cm]{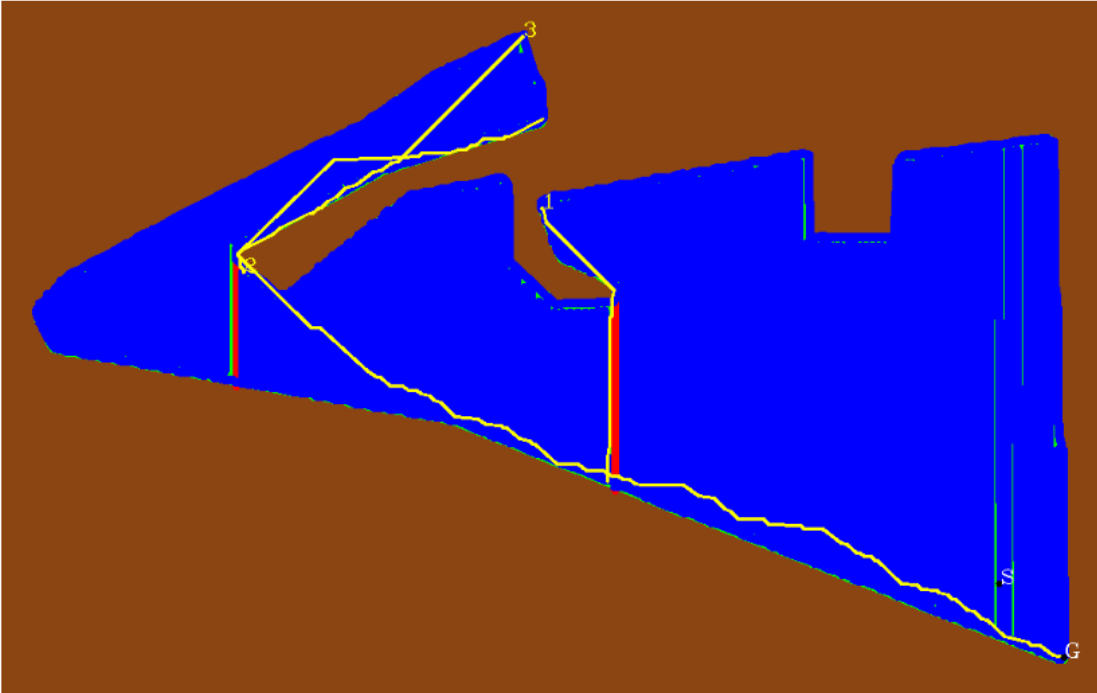} \\
            Angle & 0° & 0° & 90° & 90° \\
            Number of Decompositions & 6 & 4 & 5 & \textbf{3} \\
            Mowing Distance & 2264.8m & 2384.5m & 2242.3m & \textbf{2241.1m} \\
            Non Mowing Distance & 114.8m & 148.4m & 125.7m & \textbf{107.2m} \\
            Coverage \% & 97.69 & 98.51 & \textbf{99.45} & 98.80 \\
            Distance per Coverage & 24.39 & 25.69 & \textbf{23.85} & 23.85 \\
            \midrule
            Results \#3 & \includegraphics[width=3.0cm]{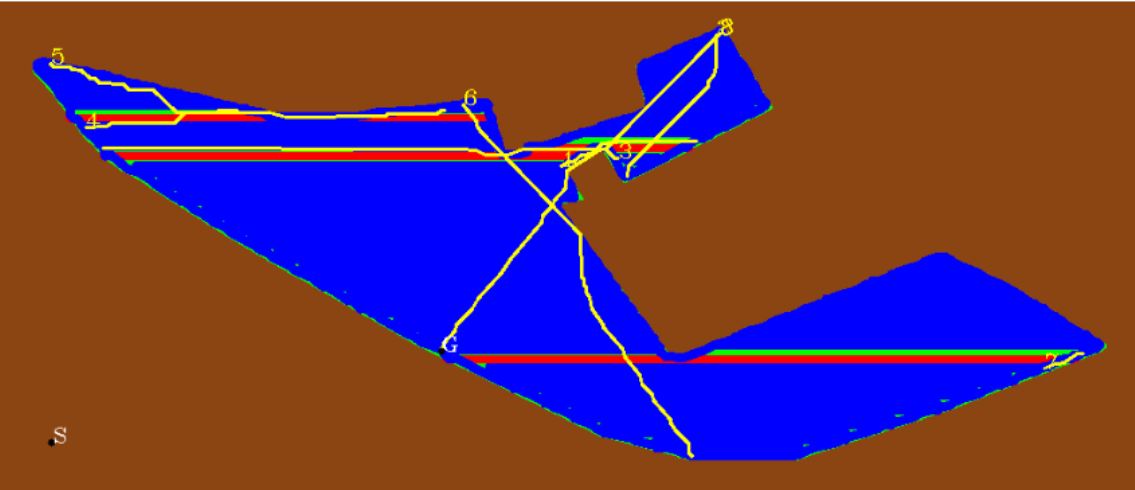} & \includegraphics[width=3.0cm]{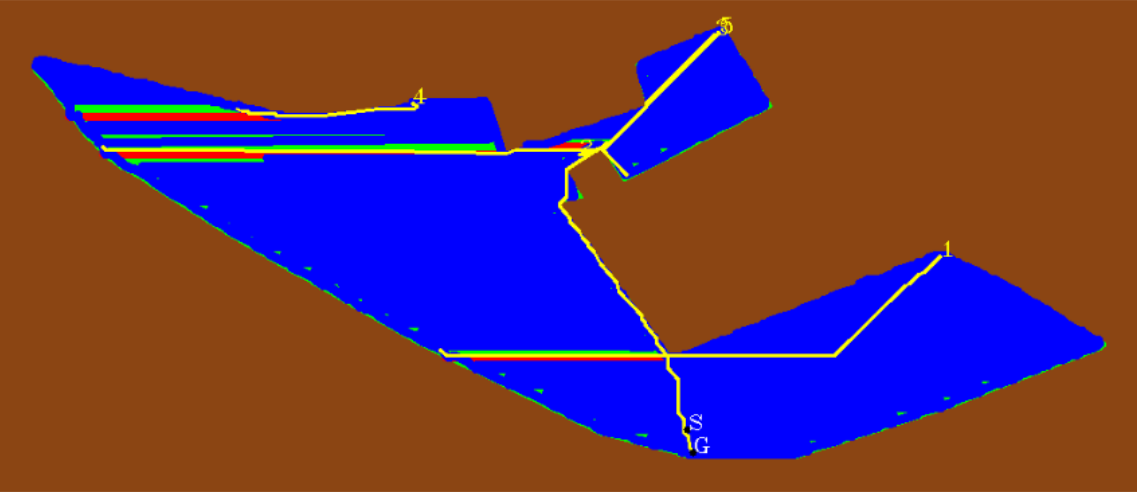} & \includegraphics[width=3.0cm]{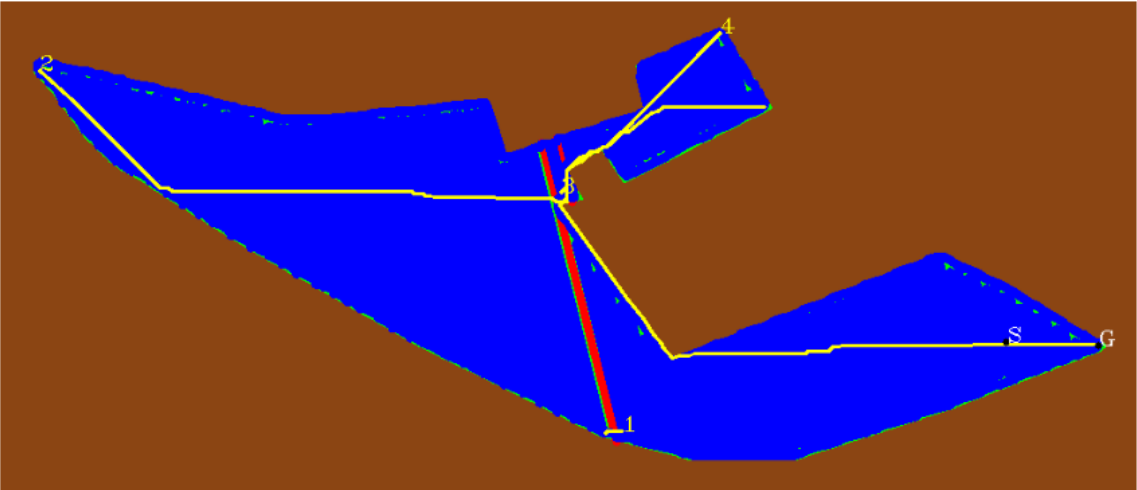} & \includegraphics[width=3.0cm]{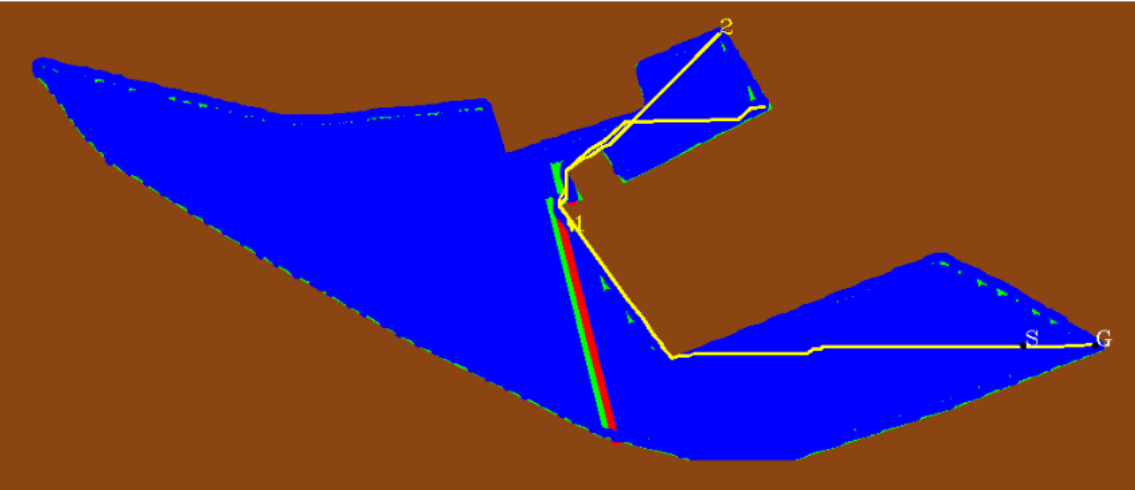} \\
            Angle & 0° & 0° & 104° & 104° \\
            Number of Decompositions & 8 & 5 & 4 & \textbf{2} \\
            Mowing Distance & 1192.6m & 1272.5m & 1190.0m & \textbf{1160.4m} \\
            Non Mowing Distance & 123.0m & 98.5m & 87.9m & \textbf{59.6m} \\
            Coverage \% & 96.06 & 97.83 & \textbf{98.56} & 97.95 \\
            Distance per Coverage & 13.71 & 13.99 & 12.97 & \textbf{12.43} \\
            \midrule
            Results \#4 & \includegraphics[width=3.0cm]{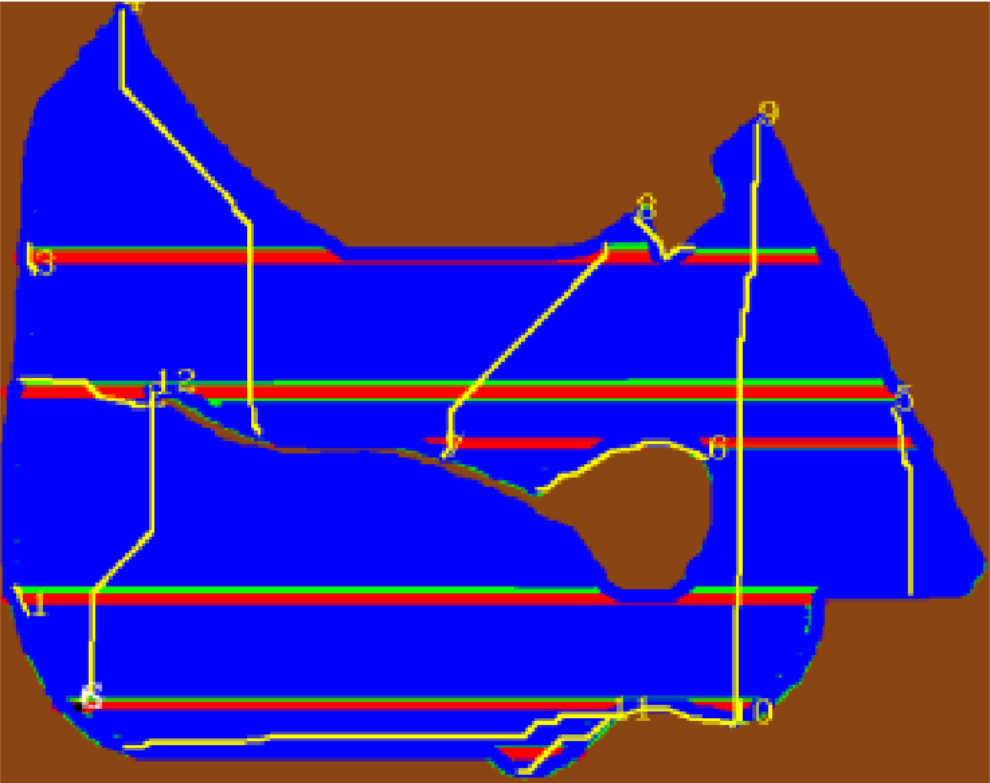} & \includegraphics[width=3.0cm]{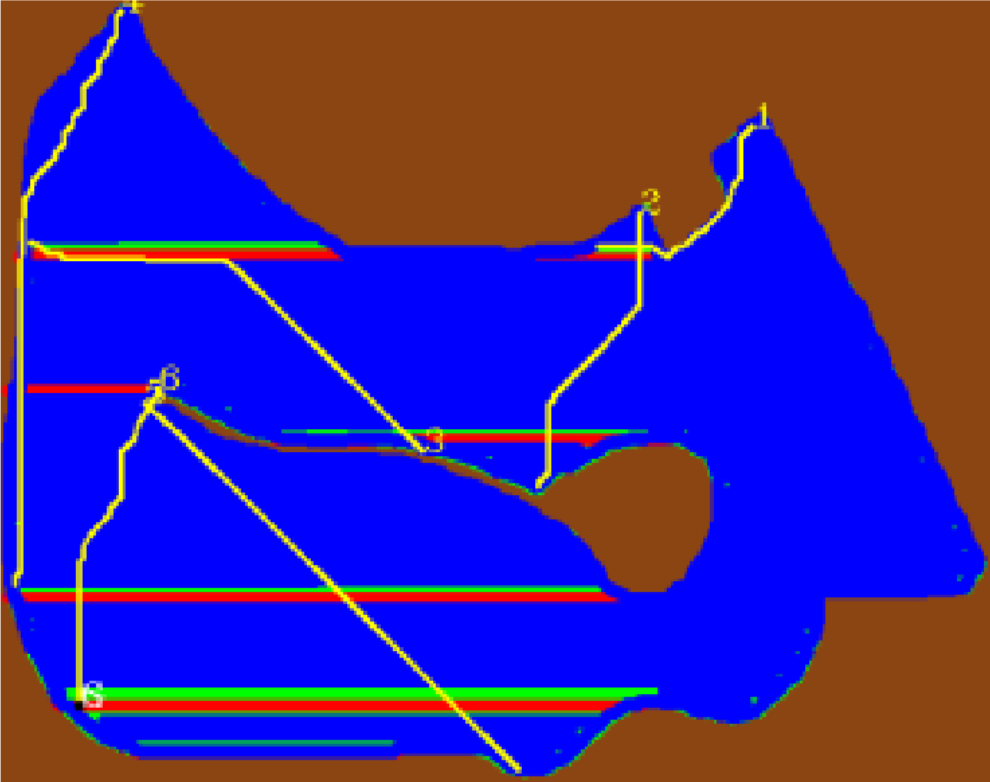} & \includegraphics[width=3.0cm]{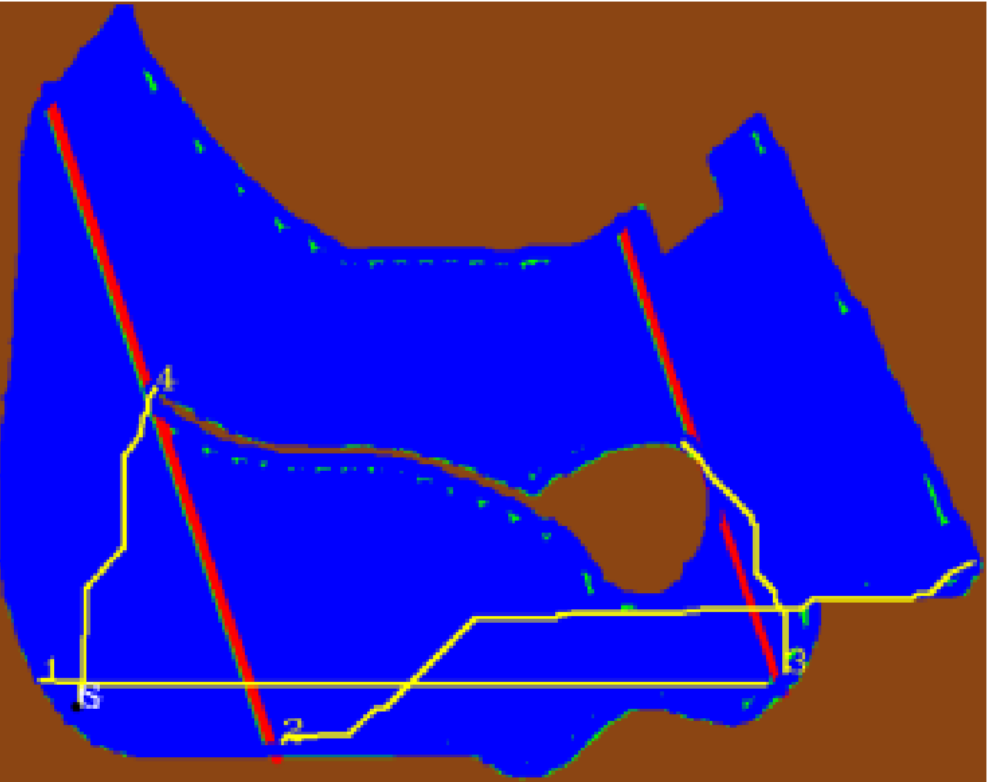} & \includegraphics[width=3.0cm]{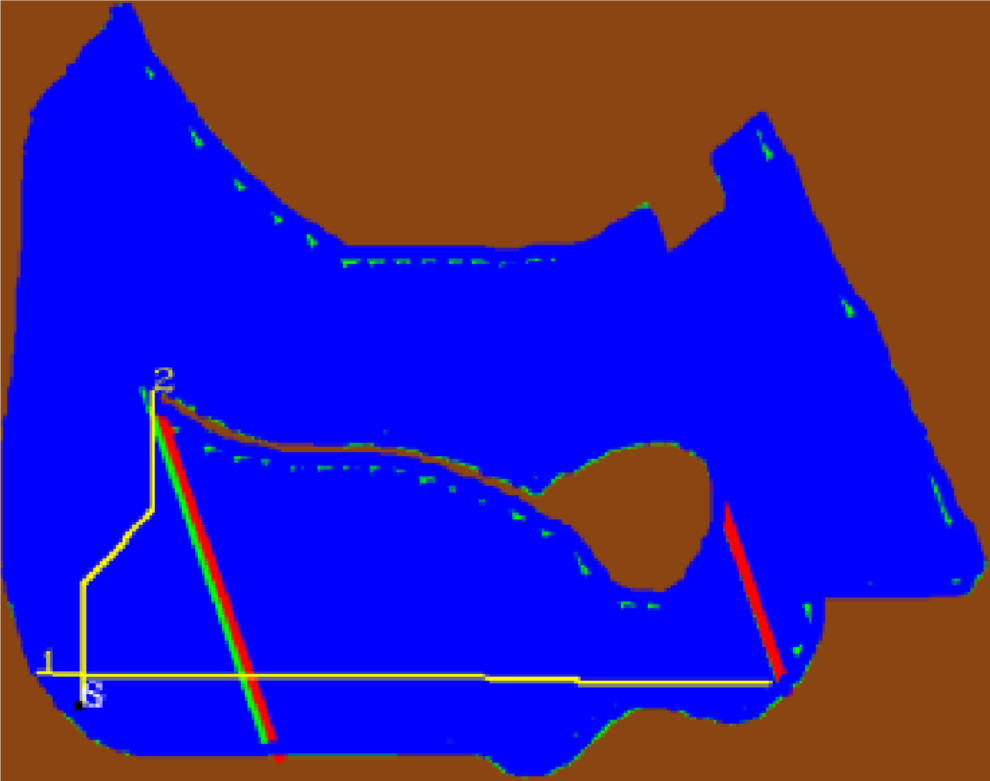} \\
            Angle & 0° & 0° & 109° & 109° \\
            Number of Decompositions & 12 & 6 & 4 & \textbf{2} \\
            Mowing Distance & \textbf{1232.0m} & 1322.2m & 1304.5m & 1336.6m \\
            Non Mowing Distance & 99.6m & 101.3m & 73.4m & \textbf{64.7m} \\
            Coverage \% & 95.56 & 95.92 & \textbf{98.76} & 98.60 \\
            Distance per Coverage & \textbf{13.94} & 14.88 & 13.96 & 14.26 \\
            \bottomrule
\end{tabular}
\caption{Summary of the algorithm test results. Each column represents a combination of the two parameters of interest: whether decomposed sections are merged and the angles of decomposition - compared between the reference 0° (horizontal) angle and the optimal angle (angle with minimum subsections). Yellow lines indicate the non-mowing path. Red lines represent subsection boundaries. Generated coverage paths are expanded to the size of the mowing width to visualize accurate coverage.}
    \label{fig:results_table}
\end{table*}

The results table (Table \ref{fig:results_table}) illustrates how different strategies, such as merging and varying decomposition angles, affect key evaluation metrics such as coverage percentage, nonmowing length, and distance per coverage.

\noindent
 \textbf{Coverage Percentage}: 
 Generally, merging sections tend to increase the coverage percentage by reducing gaps between mowing paths. For example, in Result \#1 at the reference angle of $0^\circ$, the coverage improves from $95.18\%$ without merging to $97.61\%$ with merging. However, in Results \#3 and \#4, optimizing the decomposition angle without merging yields the highest coverage percentages—$98.56\%$ and $98.76\%$, respectively. This indicates that angle optimization alone can sometimes provide superior coverage by aligning the mowing paths more effectively with the area's geometry. The slight decrease in coverage when merging is applied in these cases may be due to the complexity introduced by merging sections, which could lead to less optimal path overlaps or increased difficulty in covering irregular areas fully.


   \noindent \textbf{Non-Mowing Distance}: Merging sections and optimizing the decomposition angle usually reduces non-mowing distance across all maps, as it minimizes unnecessary movements between separate subsections. For example, in Result \#3 at the optimized angle of $104^\circ$, non-mowing distance decreases from $87.9 \, \text{m}$ without merging to $59.6 \, \text{m}$ with merging. This highlights that merging effectively generates more continuous paths, thus enhancing operational efficiency by reducing idle travel time.

  \noindent \textbf{Distance per Coverage (DC)}:  Angle optimization consistently reduces this DC metric, improving efficiency. For instance, in Result \#1, optimizing the angle from $0^\circ$ to $36^\circ$, without merging, decreases the DC value from $21.27$ to $20.03$; whereas with merging the DC value decreases from $21.62$ to $20.65$. However, exceptions exist; in Result \#4 without merging scenario, the lower DC value of $13.94$ occurs at the reference angle of $0^\circ$, but also has a lower coverage percentage of $95.56\%$. This suggests a trade-off between efficiency and completeness—while the mower travels less distance per unit percentage area, it covers less of the total area. Therefore, when evaluating mowing strategies, it's important to balance Distance per Coverage with the coverage percentage to ensure both efficient operation and comprehensive area coverage.

These results show that merging and angle optimization help in optimizing mowing paths which results in efficient coverage with minimal energy usage. Our algorithm is designed to directly optimize decomposition angles and merging leading to more efficient mowing. 

\begin{figure*}[!t]
    \centering

    \begin{subfigure}{\textwidth}
        \centering
        \captionsetup{labelformat=empty} 
        \begin{tabular}{c c c}
            \begin{subfigure}[t]{0.3\textwidth}
                \centering
                \includegraphics[width=\textwidth,height=5cm,keepaspectratio=false]{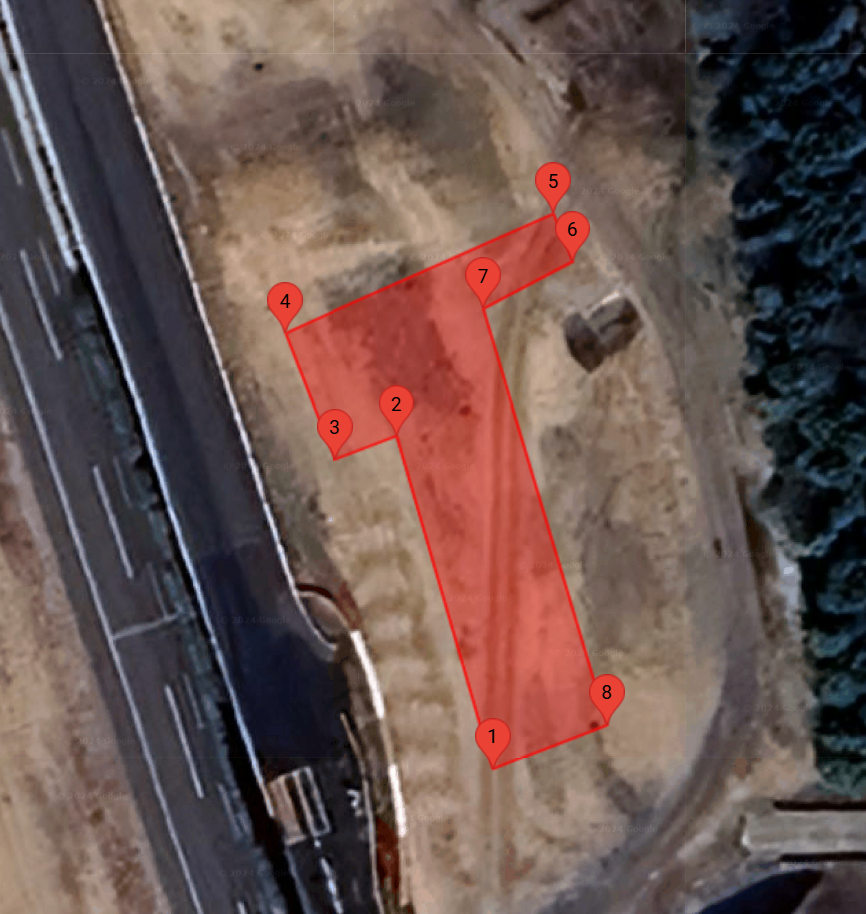}
                
                \label{fig:user_polygon_input_2}
            \end{subfigure} &
            \begin{subfigure}[t]{0.3\textwidth}
                \centering
                \includegraphics[width=\textwidth,height=5cm,keepaspectratio=false]{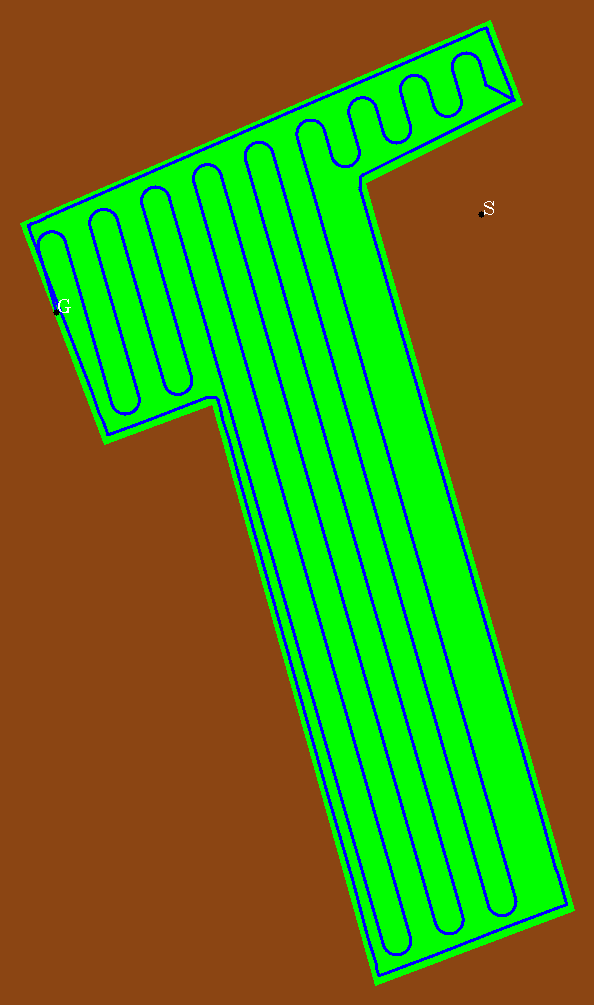}
                \label{fig:simulator_preview_2}
            \end{subfigure} &
            \begin{subfigure}[t]{0.3\textwidth}
                \centering
                \includegraphics[width=\textwidth,height=5cm,keepaspectratio=false]{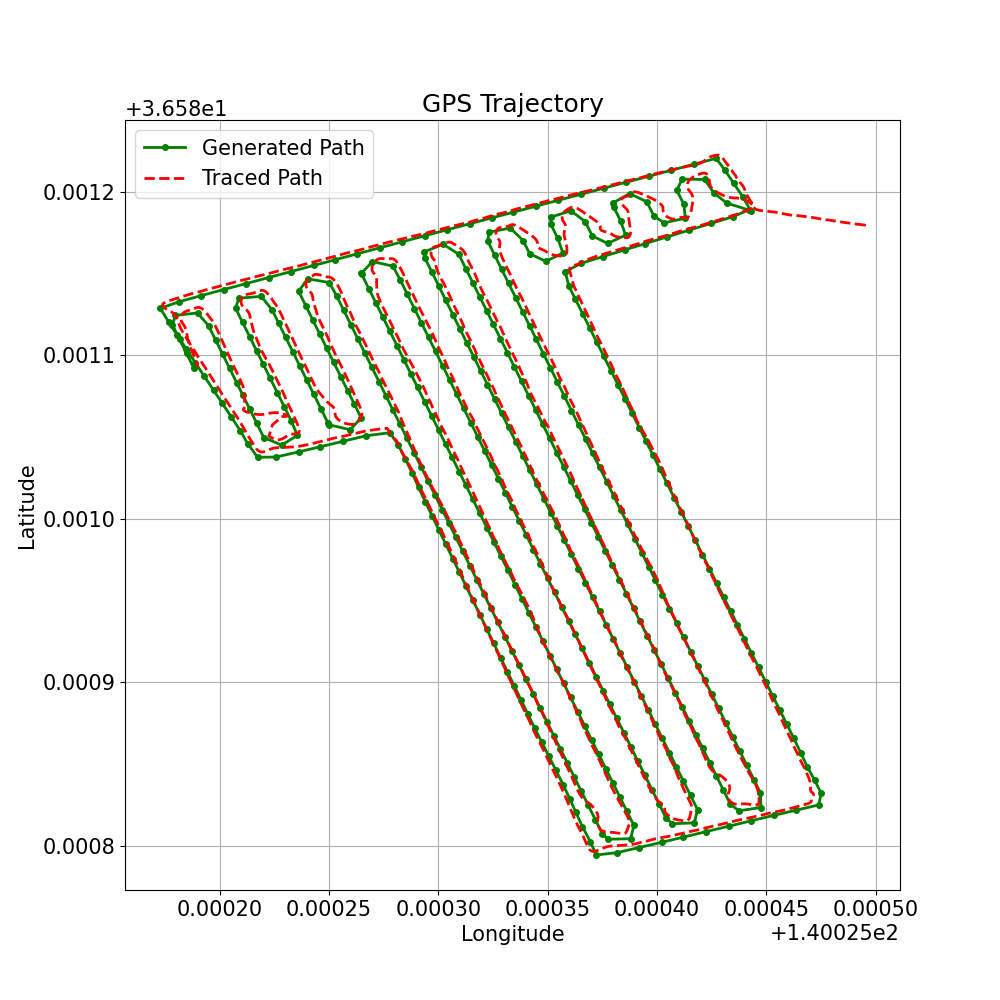}
                \label{fig:gps_output_plot_2}
            \end{subfigure}
        \end{tabular}
        
    \end{subfigure}
    
    \vspace{1em}
    
    \begin{subfigure}{\textwidth}
        \centering
        \begin{tabular}{c c c}
            \begin{subfigure}[t]{0.3\textwidth}
                \centering
                \includegraphics[width=\textwidth,height=5cm,keepaspectratio=false]{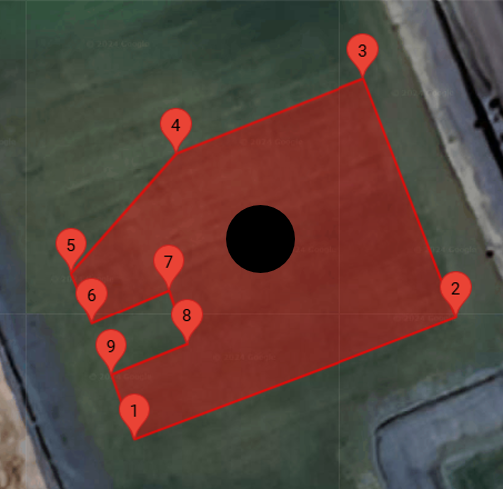}
                
                \caption{User-defined polygon on  aerial map}
                
                \label{fig:user_polygon_input_3}
            \end{subfigure} &
            \begin{subfigure}[t]{0.3\textwidth}
                \centering
                
                \includegraphics[width=\textwidth,height=5cm,keepaspectratio=false]{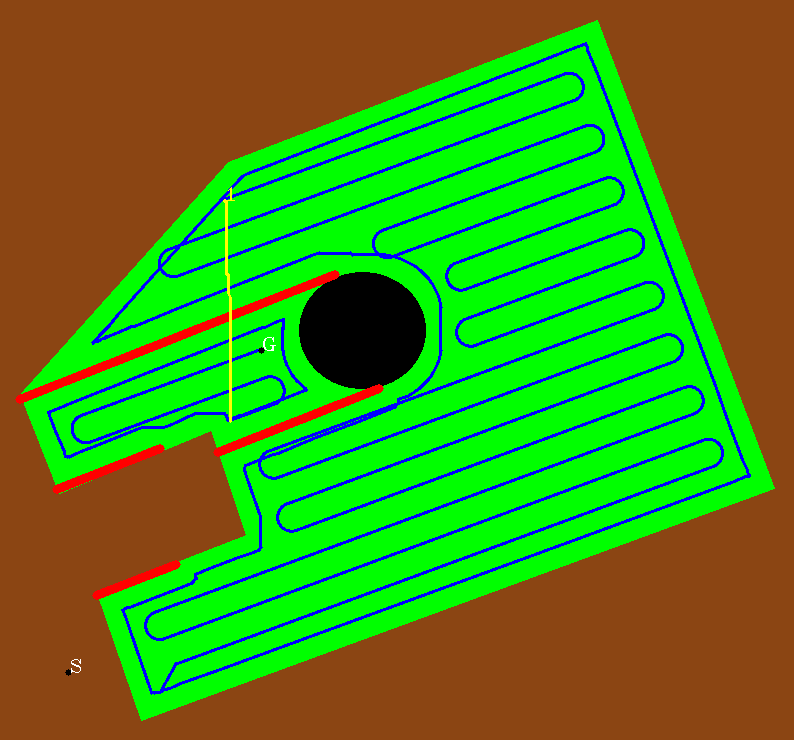}
                
                \caption{Generated path visualization}
                \label{fig:simulator_preview_3}
            \end{subfigure} &
            \begin{subfigure}[t]{0.3\textwidth}
                \centering
                \includegraphics[width=\textwidth,height=5cm,keepaspectratio=false]{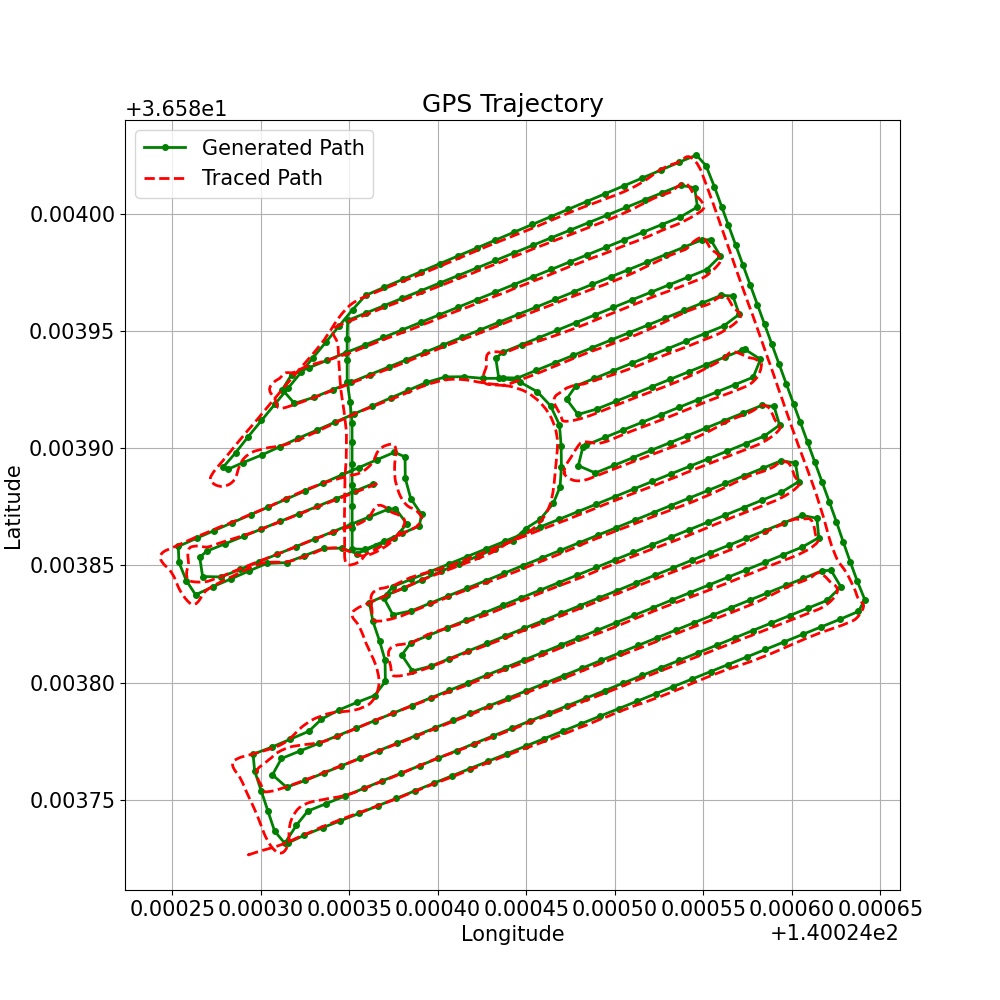}
                
                \caption{Generated vs traced GPS trajectory}
                
                \label{fig:gps_output_plot_3}
            \end{subfigure}
        \end{tabular}
        \captionsetup{labelformat=empty} 
        
    \end{subfigure}
    
    \caption{Pipeline validation for actual mowing task. (a) A user-defined boundary polygon overlaid on an aerial map. (b) A simulator preview of the generated coverage path, including offset boundaries and turning maneuvers. (c) A comparison of the planned GPS trajectory (green) and the actual path traced by the lawnmower (red), confirming successful execution in real-world conditions.}
    \label{fig:pipeline_validation}
\end{figure*}

\subsection{Framework Validation using Hardware}

To validate the pipeline, we conducted a hardware test using a medium-sized lawnmower. Validation begins with a user-defined input on a real aerial map (refer to Fig. \ref{fig:user_polygon_input_3}). The pipeline then processes this input to generate optimized coverage paths within a simulator environment, as depicted in Fig. \ref{fig:simulator_preview_3}. We also added a boundary bordering path to ensure the completeness of the task. Once we confirm the coverage output in the simulator, the software prepares the GPS waypoints for lawnmower execution. The results, shown in Fig. \ref{fig:gps_output_plot_3}, present both the generated GPS path (green) and the actual path traced by the lawnmower (red). This comparison highlights the pipeline's ability to translate user-defined inputs into optimized paths that the lawnmower can effectively follow without significant deviation, errors, or boundary violations.

The comparison of the overlap between the input map, the generated GPS trajectory, and the actual path traced by the lawnmower provides a means to validate the end-to-end framework. The localization accuracy, however, was constrained by the current system setup, which relied solely on RTK-GNSS technology. Future enhancements, such as the integration of advanced sensor fusion techniques using camera or IMU \cite{Caron2006GPS/IMU} or the adoption of higher-precision RTK-GNSS systems \cite{ng2018performance}, have the potential to significantly improve the alignment accuracy between the planned and executed paths.

The results presented in Fig. \ref{fig:pipeline_validation} also demonstrate the versatility of our approach in adapting to various environmental conditions. The top example, which is executed on slightly uneven terrain with low GPS reliability, highlights the system's robustness in difficult environments, while the bottom example illustrates the ability to effectively handle obstacles through appropriate decompositions. The entire path, including boundary bordering and section-to-section movements, is autonomously executed without interruption, since all this tasks are effectively integrated into our path generation algorithm. This validation underscores the pipeline's effectiveness in real-world scenarios, confirming that the generated paths are both practical and reliable for autonomous mowing applications.



\section{Conclusion}
We have introduced a comprehensive end-to-end CPP framework specifically designed for the operational needs of autonomous robotic lawnmowers. Central to our approach is the AdaptiveDecompositionCPP algorithm, which dynamically optimizes the decomposition angle and employs a merging strategy to minimize non-mowing distances and enhance mowing efficiency.

Experimental results, both in simulations and real-world hardware, demonstrate significant improvements in key metrics such as coverage percentage, non-mowing distance, and distance per coverage. By reducing unnecessary movements and optimizing mowing paths, our algorithm ensures efficient coverage with minimal energy usage. For instance, in comparison to standard trapezoidal and boustrophedon methods, our approach attains a higher coverage (97.2\% vs. 94.5\% and 96.8\%) while cutting non-mowing distance by up to 80\%, thereby significantly improving both efficiency and cut quality. The successful execution of generated GPS trajectories by an actual lawnmower confirms the pipeline's practical viability, accurately translating user-defined inputs into executable paths without errors or boundary violations.

The current framework is based on the assumption of a static, 2D (relatively flat) environment, which limits its ability to replan the path in the presence of moving obstacles, thereby reducing its adaptability to more dynamic environments. Additionally, there are opportunities for further optimization in the section-merging strategy, which, currently remains relatively simplistic and may result in suboptimal coverage for irregularly shaped lawns. Similarly, the pre-planning of the mowing order for sub-sections could be optimized to minimize non-mowing travel. Future advancements, including enhanced localization, multi-lawnmower coordination, and the extension of the approach to accommodate varying terrain gradients, hold the potential to significantly improve the system’s robustness and efficiency across a broader range of operational environments.

\section*{ACKNOWLEDGMENT}

We would like to acknowledge the support of Mr. Ichiro Yamaguchi for his help in supporting the development of the simulator and testing functionalities for this work.

\bibliographystyle{IEEEtran}
\bibliography{references}

\end{document}